\documentclass[final,5p,times,twocolumn]{elsarticle}

\usepackage{booktabs}

\usepackage{amsthm}
\usepackage{amsmath}
\theoremstyle{definition}

\usepackage{mathtools}

\DeclarePairedDelimiter{\floor}{\lfloor}{\rfloor}

\usepackage{color}

\usepackage[table,xcdraw]{xcolor}
\usepackage{siunitx}
\usepackage{gensymb}
\usepackage{balance}
\usepackage[utf8]{inputenc} 
\usepackage[T1]{fontenc} 

\usepackage{amssymb}

\usepackage{algorithm}
\usepackage[noend]{algpseudocode}
\makeatletter
\def\BState{\State\hskip-\ALG@thistlm}
\makeatother

\usepackage{subfig}
\usepackage{color}

\usepackage[acronym,nomain,nonumberlist]{glossaries}
\makeglossaries

\journal{Robotics and Computer-Integrated Manufacturing}
\graphicspath{{/figures/}}

\begin{document}

\begin{frontmatter}



\title{Human Intention Estimation based on Hidden Markov Model Motion Validation for\\Safe Flexible Robotized Warehouses\\
}


  \author[label1]{Tomislav Petković}
  \author[label2]{David Puljiz}
  \author[label1]{Ivan Marković}
   \author[label2]{Bj\"orn Hein}
  \address[label1]{University of Zagreb Faculty of Electrical Engineering and Computing,\\
 Department of Control and Computer Engineering,\\
 Laboratory for Autonomous Systems and Mobile Robotics,\\
 Unska 3, HR-10000, Zagreb, Croatia,\\
 {\tt\small petkovic@fer.hr, ivan.markovic@fer.hr}}

  \address[label2]{Karlsruhe Institute of Technology, \\
 Institute for Anthropomatics and Robotics, \\
 Intelligent Process Automation and Robotics Lab, \\
 Kaiserstraße 12, DE-76131, Deutschland, Karlsruhe \\
 {\tt\small david.puljiz@kit.edu, bjoern.hein@kit.edu}}


\begin{abstract}
With the substantial growth of logistics businesses the need for larger warehouses and their automation arises, thus using robots as assistants to human workers is becoming a priority.
In order to operate efficiently and safely, robot assistants or the supervising system should recognize human
intentions in real-time.
Theory of mind (ToM) is an intuitive human conception of other humans' mental state, i.e., beliefs and desires, and how they cause behavior.
In this paper we propose a ToM based human intention estimation algorithm for flexible robotized warehouses.
We observe human's, i.e., worker's motion and validate it with respect to the goal locations using generalized Voronoi diagram based path planning.
These observations are then processed by the proposed hidden Markov model framework which estimates worker intentions in an online manner, capable of handling changing environments.
To test the proposed intention estimation we ran experiments in a real-world laboratory warehouse with a worker wearing Microsoft Hololens augmented reality glasses.
Furthermore, in order to demonstrate the scalability of the approach to larger warehouses, we propose to use virtual reality digital warehouse twins in order to realistically simulate worker behavior.
We conducted intention estimation experiments in the larger warehouse digital twin with up to 24 running robots.
We demonstrate that the proposed framework estimates warehouse worker intentions precisely and in the end we discuss the experimental results.
\end{abstract}


\end{frontmatter}

\section{Introduction}
\label{sec:intro}

Substantial growth of logistics business in recent years has generated the need for larger and more efficient warehouse systems.
State-of-the-art approaches in warehouse automation include solutions such as the Swisslog's CarryPick Mobile system and Amazon's Kiva system \cite{d2012guest} which use movable racks that can be lifted by a fleet of small, autonomous robots.
By bringing the product to the worker, productivity is increased by a factor of two or more, while simultaneously improving accountability and flexibility \cite{Wurman2008}.
However, current automation solutions are based on strict separation of humans and robots; the worker is not allowed to enter the shop floor during operation due to safety reasons, since a robot with rack can weigh together up to a ton.
When moving they are posing a significant risk to all humans nearby.
Therefore, when human intervention is needed in the shop floor, the whole fleet of mobile robots has to be stopped and remain stopped until the worker has again left the shop floor.
With the increasing size of warehouses, such events immensely impact operation efficiency.
Therefore, a novel integrated paradigm arises where humans and robots will work closely together and these integrated warehouse models will fundamentally change the way we use mobile robots in modern warehouses.

Besides safety as fundamental requirement in every human robot interaction (HRI) scenario, the proposed system has to take usability into account
It has to be ensured that the worker is assisted and not impeded during work.
One way of achieving this is to on the one hand ensure that robots are avoiding the area near workers and on the other hand instruct the worker to reach its goal through robot free zones and corridors.
Given that, we assert that a future warehouse system, which will have to orchestrate and coordinate human workers and robots, would significantly benefit if it were able to estimate worker's intentions correctly and control the robots accordingly, so that warehouse operation efficiency is ensured.

\begin{figure}[htb]
\centering
\includegraphics[width=\linewidth]{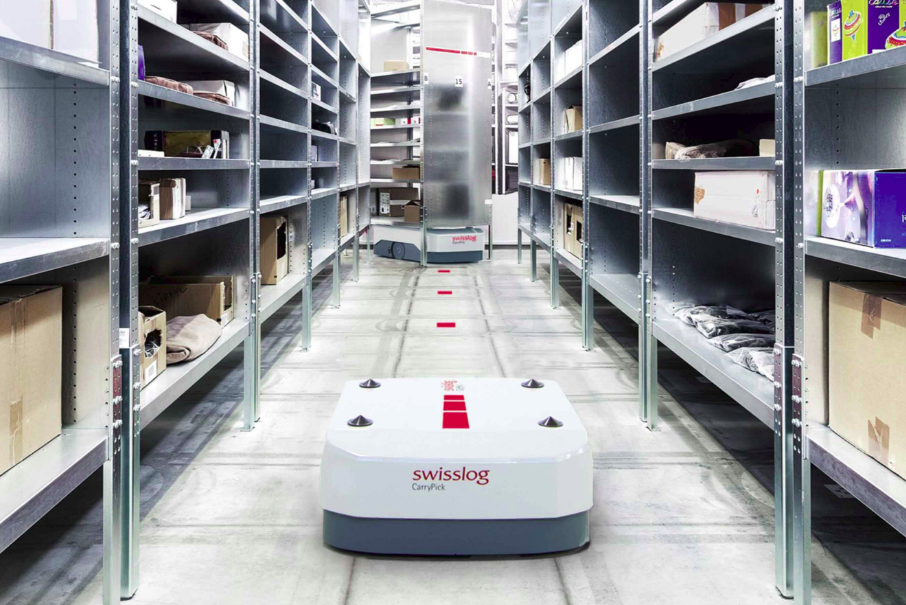}
\caption{Illustration \cite{SwissLog} of an integrated warehouse system where a fleet of mobile robots can move under the racks as well as carry them. Because of that the warehouse's layout can change and opening or closing of existing passages can occur. }
\label{fig:warehouse_example}
\end{figure}

There exists a plethora of challenges in human intention estimation, because of the subtlety and diversity of human
behaviors \cite{Bandyopadhyay2013}.
Contrary to some physical characteristics, such as the position and velocity, the human intention is not directly observable and usually needs to be estimated from human actions \cite{schlenoff2015intention}.
It is also imperative to put those actions in context because even basic behaviors, such as walking, running and jumping, are interpreted differently in, e.g., sports event, office and warehouse environment.
Furthermore, because the human intention estimation should serve as input to decision making processes \cite{Han2013}, the intentions should be estimated in real-time and overly complicated models should be avoided.
Having that in mind, only the actions with the greatest influence on intention perception should be considered as inputs to human intention recognition model.
For example, in the warehouse domain, worker's orientation and motion as well as mobile robots' movement have large effect on the goal intention recognition.
On the contrary, observing, e.g., worker's heart rate or perspiration could provide very little, if any, information on worker's intentions.  Therefore, such measurements should be avoided in order to reduce model complexity and ensure real-time operation \cite{Bandyopadhyay2013}.
In an integrated warehouse environment (Fig.~\ref{fig:warehouse_example}), worker's position and orientation, which are cues that will be used for intention estimation, need to be precisely estimated and in the present paper we assume that these quantities are readily available as well as preexisting and defined warehouse floorplan. Furthermore, given that conducting experiments in large scale operational commercial warehouses is impractical (which are usually 24h operational), methods for suitably testing the algorithms which require realistic human interaction need to be found.
For these purposes, using augmented and virtual reality seems to be a proper and practical solution.

Augmented reality (AR) and virtual reality (VR) themselves have seen a big resurgence in robotics in the recent years \cite{Williams2018VAMHRI}.
VR refers to systems where the input from the outside world is totally blocked and replaced by a system-generated input.
The first VR system was built in 1968 by Ivan Sutherland \cite{Sutherland1968Damocles}.
The device was extremely bulky and the screen resolution poor; however, it proved that VR was achievable.
The first usable VR system came in 1992 with the CAVE system \cite{CruzNeira1992CAVE}.
The CAVE is a special room where electromagnetic (now infrared) trackers track 3D glasses (now usually with active polarization) and projectors display the appropriate images on the room's surfaces.
Although viable as a system, especially for prototyping \cite{Seth2011VRPrototyping}, it is not flexible enough to be widely employed, requiring expensive sensors and a purposely designed room.
Augmented reality, in contrast to VR, seeks to add information to the input from the real world.
The first AR systems started to appear in the 1980s, mostly for military research.
The most commonly quoted first functional AR system was the \textit{Virtual Fixtures} system \cite{rosenberg1992use}, developed in 1992 by Louis Rosenberg for the US Air Force.
Quickly its usefulness became apparent in other fields as well, such as manufacturing, medicine, entertainment and robotics, where the first use cases focused on assistance in robotic teleoperation via a stereo camera pair \cite{Milgram1993HumanRobotCommunication}.
These first AR systems in robotics added data to a camera stream and displayed the enhanced view on computer monitors \cite{bischoff2004perspectives}.
Today's AR can be mostly divided into projector based, tablet/smartphone based, head mounted and computer screen based.
Given the described potential, in the present paper we will leverage these systems for providing human intention cues and construct realistic experimental scenarios.
For example, AR can be used to track worker motion inside the warehouse as well as display valuable information, e.g, navigate the worker to a specific product that needs to be picked, or assist in repairing a broken robot, while VR can be used to construct elaborate virtual warehouses with realistic simulations of worker interaction \cite{matsas2018prototyping}.

In this paper, we propose an efficient warehouse worker intention estimation algorithm for safe flexible robotized warehouses motivated by the Bayesian Theory of Mind approach.
The worker is placed inside of a warehouse with multiple potential goals.
We assume that the position and orientation of the worker are measured and that the warehouse fleet management system knows the locations of all the robots.
Based on this information, the proposed algorithm estimates the probability of the worker desiring each of the goals.
Furthermore, in our model we have also included a specific state of an \emph{irrational worker}, which could indicate to the supervisory system aberrant behavior and that intervention might be necessary.
Intentions are estimated using an hidden Markov model (HMM), where the probability of desiring each goal is modeled as a hidden state within the framework.
The HMM computes hidden state probabilities by leveraging observations generated as worker action validations based on comparing worker's motion to possible generalized Voronoi diagram paths.
Furthermore, in the present paper we also propose to use an interactive 3D simulation of a flexible warehouse in a VR environment allowing us to mimic worker's interaction and to test the intention estimation exhaustively.
Given that, we ran multiple experiments in virtual reality warehouses, as well as tested the proposed algorithm in a real-world laboratory warehouse with a worker wearing augmented reality glasses.
The laboratory warehouse as well as data and use case for real-world experiments were provided by the SafeLog project's \cite{SafeLog} partner Swisslog.

The paper is organized as follows.
In Section~\ref{sec:related} we present related research and elaborate the contributions in details.
In Section~\ref{sec:preliminaries} we give preliminaries on HMMs and Voronoi graphs.
Section~\ref{sec:methodology} presents the proposed worker intention estimation methodology.
In the end, experimental results for the real-world and virtual reality warehouses are given in Section~\ref{sec:experimental_results}, while Section~\ref{sec:conclusion} concludes the paper.

\section{Related Work}
\label{sec:related}

Our work follows up on the Bayesian Theory of Mind (BToM) framework described in \cite{Baker2014a}, where authors introduced a model for estimating hungry student's desires to eat at a particular food-truck by observing its movement.
Therein, authors argue that machines lack the Theory of Mind -- an intuitive concept humans have about other people's mental state, and propose to emulate it by an intention recognition model based on Partially observable Markov decision processes (POMDPs).
Though impressive, the BToM model does not predict the possibility of student's change of mind and does not ensure real-time operation in changing environments which is crucial in an integrated warehouse problem.
Many models addressing the problem of human intention recognition successfully emulate human social intelligence using
Markov decision processes (MDPs).
Examples of such models can be found in \cite{Bandyopadhyay2013}, where authors proposed a framework for estimating
pedestrian intention to cross the road, and in \cite{Lin2014}, where authors proposed a framework for gesture recognition and robot assisted coffee serving.

There are also works from the gaming industry perspective, proposing methods for improving the non-playable
character's assisting efficiency \cite{Nguyen2012,Fern2011}.
A driver intention recognition problem was approached in \cite{Jin2011} and intention estimation based on gaze data was introduced in \cite{10.3389/fpsyg.2015.01049}.
Both of those approaches use learning methods for training the models which has been criticized by \cite{Vasquez2009} emphasizing the drawback of using motion pattern learning techniques for trajectory estimation or intention recognition.
Authors assert that such techniques operate offline and imply that at least one example of every possible motion pattern is contained in the learning data set which does not hold in practice.
They propose using growing hidden Markov models (GHMM) for predicting human motion, a problem which we consider dual to the human intention estimation in the warehouse domain.
GHMMs can be described as a time-evolving HMMs with continuous observation variables where model structure and parameters are updated every time a new observation sequence is available.
That kind of approach can be applied to human intention recognition problem, because it enables adding new goals during the experiment as well as an elegant framework for learning the model parameters online.
Assistive technology such as smart homes \cite{Rafferty2017}, exoskeletons \cite{long2018human} and semi-autonomous wheelchairs \cite{Taha2008} benefit also from precise human intention recognition.
In \cite{Taha2008}, authors propose a POMDP driven algorithm for wheelchair control taking into account the uncertainty of  user's inputs because of, e.g., unsteady hands.
The chair predicts user's intention and autonomously enacts the intention with only minimal corrective input from the user.
The authors also suggest that humans usually focus on moving from one spatial location to another, i.e., hallway to kitchen, without worrying about the optimality of exact steps that come in between.
In the present paper, we build our model using similar assumptions about human spatial understanding.
In \cite{Han2013} Anh and Pereira offer thorough review of human intention recognition area emphasizing its potential applications in decision making theory.

Another emerging assistive technology is Mixed Reality (MR), a term usually including Augmented Reality, Virtual Reality, or a mix of thereof. As stated before one can leverage localization and sensor systems required for immersive AR for intuitive information display and the tracking of human motions, while VR can be used to realistically simulate environments for training and testing purposes. Here we give a brief overview of current MR systems and their applications.

Advancements in VR technology have seen the CAVE systems of the early 90s being replaced by cheaper and more flexible headsets such as the \textit{Oculus Rift} or the \textit{HTC Vive}.
This has sparked a boom in the field of robotics where VR has recently seen the most use as a more intuitive method for teleoperation of stationary \cite{Lipton2018VRBaxter}, mobile \cite{Jankowski2015VRMobileRobot} and humanoid robots \cite{allspaw2018remotely}.
A natural extension of such a teleoperation systems is a more immersive tool for telepresence robots \cite{Zhang2018telepresence}.
It has also emerged as a method to teach virtual robots how to preform tasks \cite{Zhang2017VRDL}, where the knowledge is then successfully transferred to a real robot.
In manufacturing \cite{Mujeber2004VRManufacturing} VR has seen use as a virtual prototyping \cite{CHOI2004VirtualPrototyping} and training tool \cite{gavish2015evaluating}.
As a training tool, VR has been shown to increase performance of trainees in other areas as well, such as medicine \cite{seymour2002virtual}, safety training in construction \cite{sacks2013construction}, and mining \cite{squelch2001virtual}.

On the AR side of things, projector based AR has seen use for visualizing robot's intentions and intuitive programming of robots in robot work cells \cite{gaschler2014intuitive}, displaying intentions of mobile robots operating in human environments \cite{Chadalavada2015thats}, as well as for debugging and rapid prototyping of robotic systems through visualization of sensor data \cite{omidshafiei2016ARPrototyping}.
Tablets have seen use in AR-assisted robot programming \cite{mateo2014hammer} as well.
Devices like the \textit{Google Tango} tablet, in addition, with inbuilt SLAM \cite{Cadena2016SLAMSurvey} can be used for markerless AR applications.
The main challenge with tablets is that it occupies the hands, preventing any work while the AR information is visualized.
With the recent releases of the ARCore and ARKit toolkits, for Android and iOS devices respectively, tablet and smartphone based AR applications are becoming an economical and straightforward interaction modality for home robots \cite{Sprute2017VirtualBorders}.
Head mounted systems can be further divided into Heads-up Display (HUD) systems and ``full-AR'' systems, with the most famous member of the former being the \textit{Google Glass} and of the later \textit{Microsoft Hololens}.
As the name implies, HUD-based systems do not have any advanced localization or computing systems and therefore are only able to display interfaces, while full AR systems are able to perform localization using SLAM and display persistent, full 3D holograms in space.
HUD systems have found applications in logistics, where they have been used for \emph{pick-by-vision} systems to quicken and ease the picking of items in warehouses \cite{Schwerdtfeger2011PickByVision}.
Since full AR systems, starting with the \textit{Hololens}, have not been on the market for long, research is just starting \cite{Williams2018VAMHRI}, with the most prominent field perhaps being AR assisted robot programming \cite{Guhl2017Hololens}.

The present paper draws upon our earlier work \cite{Petkovic2018}, where we have presented the preliminary version of the human intention estimation algorithm.
Therein, the algorithm was also motivated by the Bayesian Theory of Mind \cite{Baker2014a} and tested in a simulated $20\times 20$ cells large environment.
Worker actions were validated by comparing worker actions with the optimal policy within an MDP framework, where the value iteration algorithm needed to be executed beforehand.
In case of any changes in the environment, such as robots blocking a path that was previously free, value iteration would need to be reran in order to find the optimal policy.
Such an approach proved to be infeasible for large flexible warehouses where multiple robots could constantly block worker's path, thus changing often the configuration of the environment.
Given that, in the present paper we propose a worker intention estimation algorithm for safe flexible robotized algorithm that solves the aforementioned problem by creating first a generalized Voronoi diagram of the warehouse and running the D* algorithm in order to find the optimal path between each two nodes.
In case of a robot blocking the path, we can simply cut the edge between the nodes, thus ensuring efficient worker motion validation that is fed to the HMM as observations.
Hidden states of the HMM each model worker intentions towards the goals, additionally including a specific state which models an irrational worker.
This approach also enables us to efficiently add goals during the experiment, which was previously not possible.
In order to test the intention estimation algorithm we ran experiments in a real-world laboratory warehouse with worker wearing Microsoft Hololens augmented reality glasses.
The Hololens localization algorithm was used to generate worker location and orientation estimates.
Furthermore, in order to demonstrate the scalability of the approach to larger warehouses, we propose to use VR digital warehouse twins in order to simulate realistically worker behavior.
First, the real-world experiment warehouse was modeled in VR and the pertinent experiment was recreated in order to showcase the realistic nature of the VR experiments and compare the outputs of the intention estimation algorithm.
Then, further intention estimation experiments were ran in a larger warehouse digital twin with up to 24 running robots.
The results corroborate that the proposed framework estimates warehouse worker’s desires precisely and in an intuitive manner.

\section{Preliminaries on HMMs and Voronoi graphs}
\label{sec:preliminaries}

Markov decision processes (MDPs) constitute a mathematical framework which models a system taking a sequence of actions under uncertainty to maximize its total rewards \citep{Bandyopadhyay2013}.
More precisely, an MDP is a discrete time stochastic process represented as tuple $(\mathcal{S}, \mathcal{A}, T, \mathcal{R}, \gamma)$, where $\mathcal{S}$ is set of states and $\mathcal{A}$ is set of actions.
After an action $a \in \mathcal{A}$ is taken, system moves from the current state $s \in \mathcal{S}$  to a new state $s' \in \mathcal{S}$.
We define the conditional probability function $T(s,a,s') = p(s' | s, a)$ which gives the probability that the system lies in $s'$ after taking the action $a$ in state $s$, thus capturing system's uncertainty.
Taking an action also yields an immediate reward $R(s,a)$ and the goal of the system is to choose the sequence of actions that maximizes the expected total reward $E\big(\sum_{t=0}^{\infty}\gamma^tR(s_t,a_T)\big)$.
To prevent an infinite-horizon case where all positive rewards sum to infinity \cite{Thrun1999}, one uses a discount factor $\gamma \in (0,1)$ which reflects system preference of immediate rewards over future ones.
The MDP model can be too restrictive to be applicable to many problems of interest \cite{Rabiner1989a} because it assumes that all states are fully observable.

The hidden Markov model (HMM) is an MDP extension including the case where the observation is a probabilistic function of the state, i.e., the resulting model is a doubly embedded stochastic process with an underlying process that is not observable (it is hidden), but can only be observed through another set of stochastic processes that produce the sequence of observation.
In general, when using HMMs we are interested in solving one of the following three problems.
First, given an existing HMM and an observed sequence, we want to know the probability that the HMM could generate such a sequence (the scoring problem).
Second, we want to know the optimal state sequence that the HMM would use to generate the sequence of such observations (the alignment problem).
Third, given a large amount of data, we want to find the structure and parameters of the HMM which best account for the data (the training problem).
In this paper we focus on optimal state sequence problem, i.e., the alignment problem, for warehouse worker actions with the aim of estimating worker intentions.

The HMM gives us a procedure for handling observations and turning them into intention estimations.
In order to produce observations we need to also find a way to efficiently encode the environment and validate worker's actions within.
For this task we selected the Voronoi diagram.
Voronoi diagram partitions a plane into regions based on distance to predefined points.
The idea is that for each predefined point a corresponding region consisting of all the points closer to that point than to any other predefined point is found.
More formally, the definition of Voronoi region for point $p_i \in P$ is given by:
\begin{equation}
  V(p_i) = \{\vec{x}\;	 \big\rvert \;||\,\vec{x}-\vec{p_i}\,|| \leq ||\,\vec{x}-\vec{p_j}\,||, \forall_{j \in \{1 \dots |P| \}} i \neq j \},
\end{equation}
where $|| \cdot ||$ is usually the Euclidean distance \cite{Aurenhammer1991}.
We call the set given by $V = \{ V(p_1), V(p_2), ... V(p_n)\}$ the Voronoi diagram of $P$.
The question is, how to construct a Voronoi diagram in practice?
The generalized Voronoi diagram (GVD) is a discrete form of the Voronoi diagram defined as the set of points in free space to which the two closest obstacles have the same distance \cite{Choset2000}.
In mobile robotics applications, the GVD can be constructed from an occupancy grid map of the environment which is usually obtained by mapping the environment with a mobile robot \cite{Thrun2003} or by parsing the existing floorplans.
In the present paper, we will use the latter approach, since we have warehouse plans at our disposal.

\section{Worker intention estimation method}
\label{sec:methodology}

Theory of Mind (ToM) is the intuitive grasp that humans have of their own and other people's mental states, how they are structured, how they relate to the world, and how they cause behavior \citep{Baker2014a}.
Human beings understand that others have certain desires and that those desires guide them to act using means most likely to achieve them.
However, \emph{explanation by rationalization} reasoning, which ToM assumes, is highly contextually dependent and translating such causal behavior model to machines is not an easy task.
Having that in mind, we limit the proposed model to a problem of estimating intention of a human worker in a highly flexible robotized warehouse.
The worker can perform tasks such as maintaining the robots or picking the items from the racks containing goods.
We assume that there is finite number of possible goal locations which are usually in front of the \emph{interesting} racks and that at least one goal is known before the start of the experiment.
Furthermore, we also assume that the position and orientation of the worker are measured.
This can be achieved by, e.g., augmented reality glasses, other types of wearable sensors, or special vests equipped with vision sensors as developed in the scope of the project SafeLog \cite{SafeLog}.
Dennett \cite{dennett1989intentional} has proposed that social reasoning abilities rely on \emph{intentional stance}, i.e., the assumption that agents have beliefs about the world and their situation in it and will behave rationally to achieve their desires.
We argue that, in the warehouse domain, rationally behaving with respect to the desired goal manifests as moving towards that goal's location, and that since our agents are workers, trained professionals, they are highly likely to behave rationally within this context.

In order to determine the most likely goal the worker is moving to, we need to apply a complete and globally optimal
path planning algorithms and compare the worker's motion on-line with the algorithm output (details are discussed later).
We assert that worker following approximately a globally optimal path is a reasonable assumption, since the worker is acquainted with the warehouse layout and will plan its motion in accordance to it.
Given that, we find that for the problem at hand, where action uncertainties are reduced, frameworks such as POMDPs used in \cite{Baker2014a} are not necessary.
Moreover, the planning algorithm must be able to quickly replan the path with the appearance of moving obstacles such as mobile robots carrying the racks.
Planning using MDP solvers \cite{Thrun1999} offers a well grounded tool for human intention recognition by allowing elegant comparison of agent actions with respect to the optimal policy \cite{Petkovic2018}; however, it lacks the ability to quickly replan, since the optimal policy for efficient intention recognition must be computed offline for realistic warehouses.
Given that, in the present paper we choose to use the D$^*$ algorithm \cite{Stentz1997} for finding the globally optimal path to the goals.
However, having in mind that the modern warehouses are growing in size we aim to reduce the complexity of mentioned search problem.
One approach to alleviating the complexity is to reduce the precision of the warehouse occupancy grid representation, thus reducing the D$^*$ algorithm search space; however, we assert that it could jeopardize the proposed human intention recognition performance and cannot be applied to arbitrary large warehouse, thus directly impacting the proposed algorithm's ability to generalize.
\begin{figure}[!t]
\centering
\includegraphics[width=\linewidth]{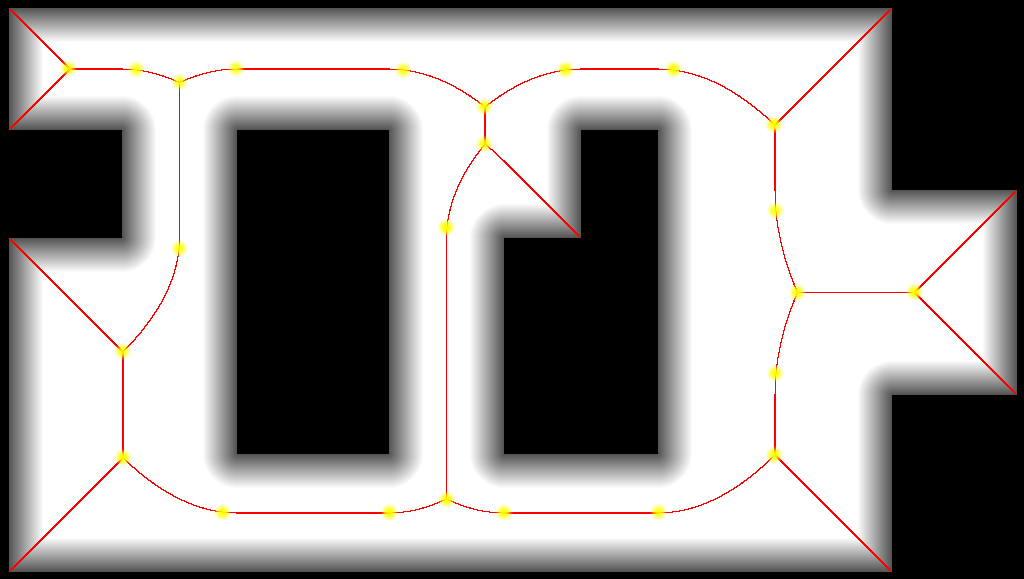}
\caption{Generalized Voronoi Diagram (red) of Swisslog's warehouse in Ettlingen with highlighted graph's nodes (yellow). Untraversable parts of the warehouse such as walls or racks are denoted with black color. Shades of grey denote the distance of the traversable part of the warehouse from the obstacles. The distance is used for GVD generation. }
\label{fig:voronoi_example}
\end{figure}

To solve the aforementioned challenges, we propose to use generalized Voronoi diagrams (GVDs) for reducing the search space without losing valuable precision for intention recognition.
Motivation for using GVDs in our work is manifold.
First, partitioning the plane using GVDs allows us to limit the search space on the Voronoi graph nodes, thus greatly reducing the search time.
Second, moving along the edges of a Voronoi graph ensures the greatest possible clearance when passing between obstacles.
This property resembles assumed human path planning in a warehouse application, because human beings are generally not prone to walking in the proximity of warehouse racks (note that in our example robots can also pass under the racks).
Finally, moving obstacles, such as mobile robots, in flexible warehouse systems can obstruct worker's path necessitating replanning of the optimal path towards that goal.
The replanning can easily be achieved using graph search algorithms by discarding the edge of the Voronoi graph the robot is currently occupying.
On the other hand, we can imagine a scenario where new possible passages could appear.
This could happen when a mobile robot takes a rack and frees up space in the middle of the rack block.
In order to handle that event, a new GVD would have to be generated from the occupancy grid map of the warehouse.
This could impede the online application of the proposed approach; however, in order for that to happen, since usually one rack block has two columns of racks, multiple racks close to each other would have to be carried away at similar time intervals.
Furthermore, we assert that workers in robotized warehouses would generally not be allowed to use such passages for safety reasons.
Having that in mind, from the methodological perspective for the approach, we propose to use the D$^*$ algorithm on GVD nodes for worker path planning.
In the sequel we describe the proposed human intention recognition algorithm in details and divide it in three parts: creation of the warehouse GVD (offline), worker action validation, and worker intention estimation (both online).

\subsection{Creating warehouse generalized Voronoi diagram}\label{offline}

We have already emphasized the necessity of performing human intention recognition online.
In order to ensure online operation, we need to perform time consuming parts of the algorithm before the start of the experiment.
First, we generate the GVD \cite{Lau2013} of the warehouse using its floorplan.
Example of such a graph generated on Swisslog's warehouse using an occupancy grid with cell size of $5$\,mm can be seen in Fig.~\ref{fig:voronoi_example}.
After generating GVD, we select all graph nodes for further processing.
It is worth noting that it is possible to insert additional nodes at arbitrary locations, but erasing generated ones  except dead ends, is not allowed because it would impede graph's connectivity.
Also, at least one goal node must be added before the experiment starts, since it is not possible to emulate ToM inference without having any hypotheses about possible goal locations.
In this work we add only the goal nodes and discard the dead ends creating a node set which we denote $\mathcal{N}$.
With the obtained node set, we run the D$^*$ algorithm to find the optimal path between each two nodes and save all the relative distances in a matrix \textbf{F}, where element $\textbf{F}_{i,j}$ denotes distance in pixels between nodes $i$ and $j$.
This might not be the optimal approach to finding relative distances between the nodes, but this part if performed offline and done only once before the start of the experiment.

\subsection{Worker action validation}
\label{online}

During the online phase we monitor worker's position and orientation provided by the Microsoft HoloLens augmented reality device as well as  positions of mobile robots (provided by the warehouse fleet management system).
Mobile robots are treated as moving obstacles with radius $r=1$\,m and the worker's weareble device shows positions of nearby robots.
If the robot is located on a GVD edge, we cut that edge from the graph and update the relative distance matrix \textbf{F} using the D$^*$ algorithm.

\begin{algorithm}[!t]
\caption{Human intention recognition}\label{hir}
\begin{algorithmic}[1]
\While{True}
\If{$\text{Worker moved or turned significantly}$}
\State $d \gets \text{Modulated distance to every goal}$
\For{$\text{Proximate positions and orientations}$}
\State $D_i \gets \text{Modulated distance to every goal}$
\EndFor
\State $\text{Update intention estimation(d, Di)}$
\EndIf
\EndWhile
\end{algorithmic}
\end{algorithm}

\begin{figure}[!t]
\centering
\includegraphics[width=.95\linewidth]{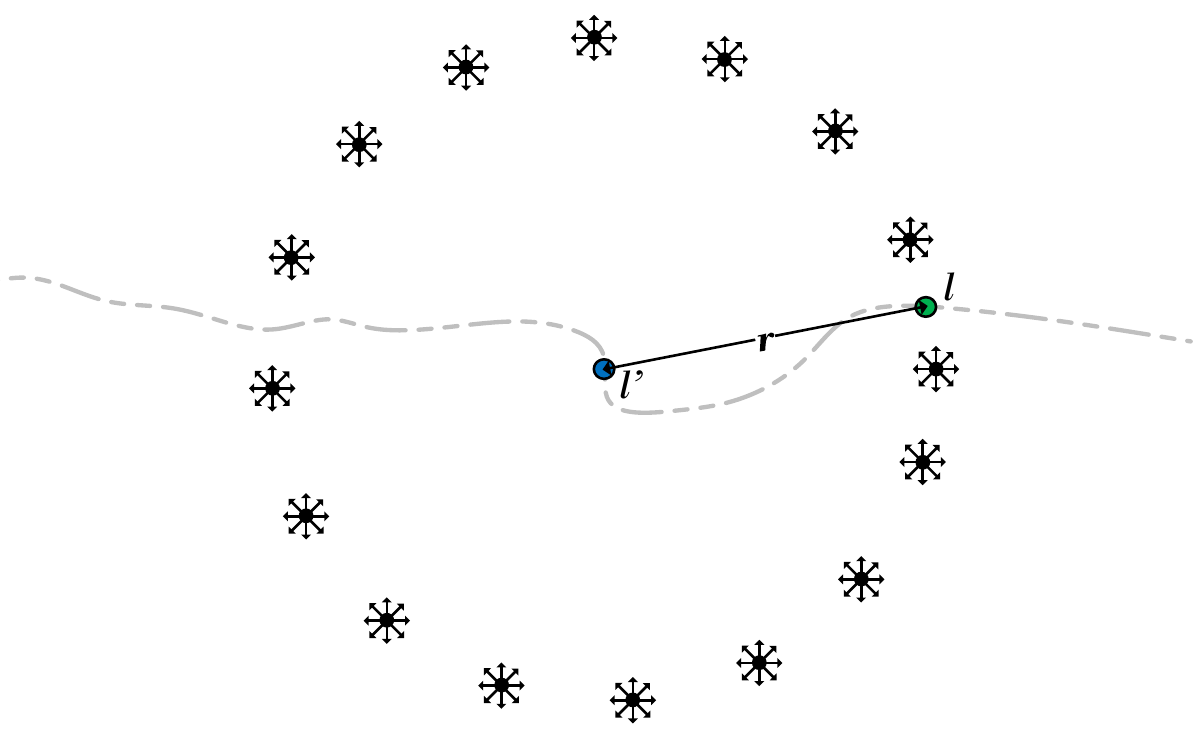}
\caption{Illustration of generating additional points $p$ with orientations $\theta$. We validate the worker's motion by comparing the vector  $\textbf{c}$ of the worker's position and orientation at location $l$ with vectors $\textbf{c}$ of the newly generated points. }
\label{fig:v_vector_illustration}
\end{figure}

The final objective of the human intention algorithm is to estimate towards which goal the worker is currently going to by observing worker motion and comparing it to the optimal path to each of the goals.
With each worker's position and orientation information update, we check if (i) worker's position mapped to the occupancy grid floor plan has changed or if (ii) worker's orientation has changed more than $\frac{\pi}{8}$ from the last intention estimation update.
If any of these conditions is met, we perform an intention estimation update by associating observed worker's position and orientation with each of the nodes in set $\mathcal{N}$ using vector $\textbf{c}$ as follows:
\begin{equation}
c_i =
\begin{dcases}
0, & \text{an unobstructed straight line exists} \\
 & \text{between the worker and the \textit{i}-th node}\\
\mathcal{G}(d_i, \sigma^{2})\cdot\Phi(\theta_i), & \text{otherwise.}
\end{dcases}
\label{eq:conditional}
\end{equation}
In \eqref{eq:conditional} $\mathcal{G}$ is a Gaussian function with variance $\sigma^2=0.005$, which we obtained experimentally, and $d_i$ is distance between worker's position and $i$-th node's location.
Bell shaped functions such as Gaussians have been used for navigating in continuous spaces \cite{Hengster-Movri??2010}, which motivated us to choose them as a proximity measure.
They are smooth and monotonically decreasing functions of distance and have a non-zero value on the entire domain, allowing an intuitive association of worker's position with every visible node.
We also assert that every worker's position will always be associated with at least one node because of GVD's space covering properties.
The only exception to forementioned claim is if the worker is trapped by mobile robots but we argue that warehouse management system must never allow such event to occur for obvious safety concerns.
Furthermore, we also modulate the Gaussian with the following triangular function:
\begin{equation}
\Phi(\theta_i) = \frac{\pi-|\theta_i|}{\pi^2},
\label{eq:triangular}
\end{equation}
where $\theta_i \in [-\pi, \pi]$ is the difference between worker orientation and the angle at which the worker sees $i$-th node.
It amplifies the association with those nodes the worker is oriented at, since we assume that the worker will look at the path it is planning to take \cite{Petkovic2018}.
We also need to define the isolation matrix $\textbf{I}^{n \times g}$, where $n$ is the total number of nodes and $g$ is number of goal nodes:
\begin{equation}
\textbf{I}_{i,j} = \begin{dcases}
1, &  i=n-g+j\\
0 & \text{otherwise.}
\end{dcases}
\label{eq:isolation}
\end{equation}
We normalize the association vector $\textbf{c}$ and obtain modulated approximate distance vector $\textbf{d}$ by multiplying it further with the distance matrix $\textbf{F}$ and isolation matrix $\textbf{I}$:
\begin{equation}
\textbf{d} = \textbf{c} \textbf{F}  \textbf{I}.
\label{eq:distance}
\end{equation}
Each element of vector $\textbf{d}$ represents modulated measure of distance to the respective goal.
In order to find out if the worker is moving towards or away from the goal, we compare this distance to alternative worker positions and orientations.
We take the location $l'$ at which the worker was prior to the last intention estimation update, and calculate the difference $r$ between that position and current worker's position $l$.
Then, we generate set of $m=16$ equidistant points $p$ on a circle around $l'$ with the radius $r$ which is shown in Fig.~\ref{fig:v_vector_illustration}.
For each point $p_i \in p$ and potential worker orientation $\theta \in \{-\pi, -\frac{3\pi}{4}, ... \hspace{3pt}, \pi\}$ we repeat the calculation of vectors $\textbf{c}$ and $\textbf{d}$ and append the result to potential modulated distances matrix $\mathbf{D}$.
Computing the matrix \textbf{D} enables us to validate worker's motion with respect to states it \textit{could be in}, rather than to the state it \textit{had been to}.
The proposed algorithm pseudocode can be seen in Algorithm~\ref{hir}.

\begin{figure}[!t]
\centering
\includegraphics[width=1\linewidth]{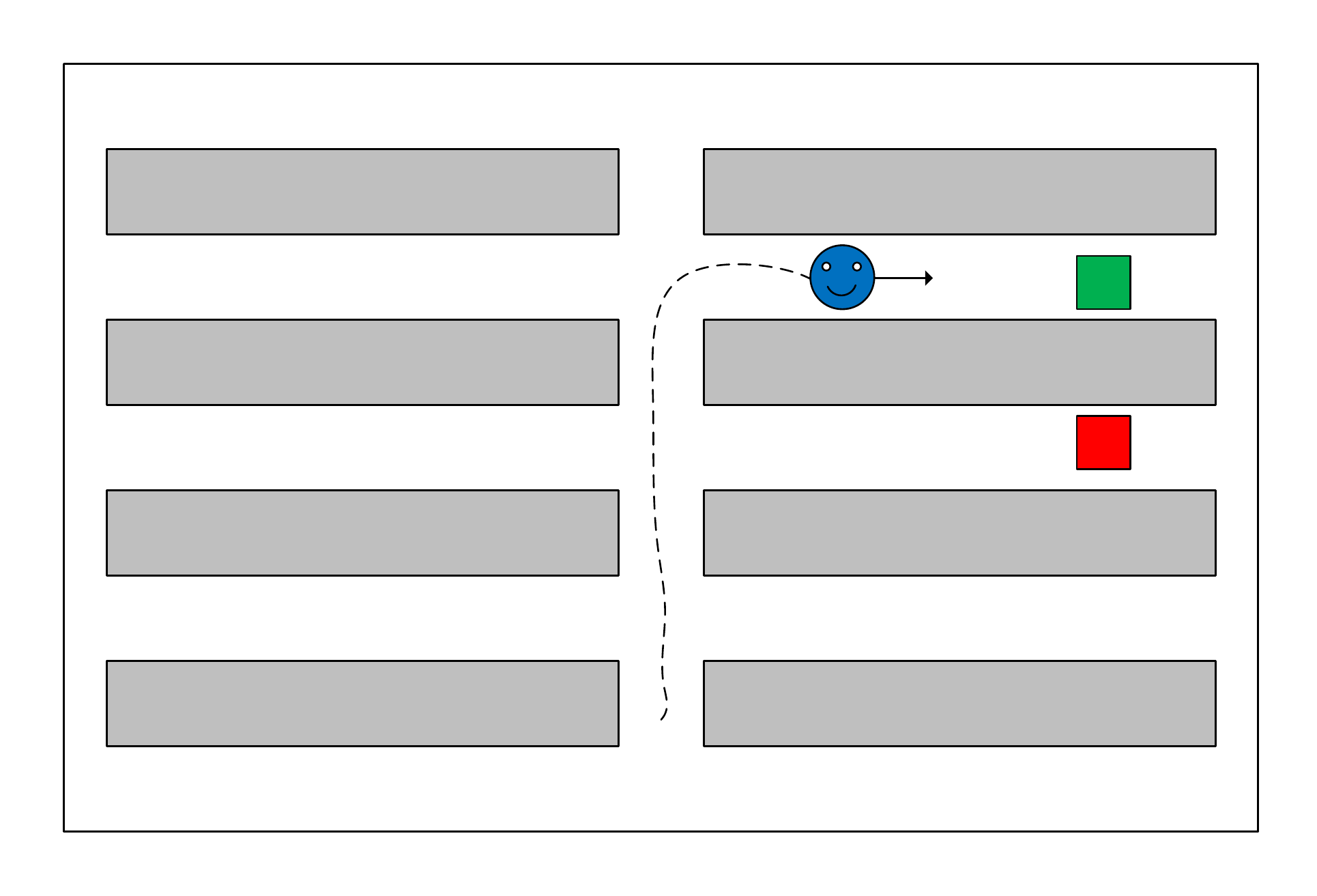}
\caption{The warehouse worker (blue circle) has previously decided not to turn towards the goal labeled with the red box. It is intuitive that the worker desires the green goal more than the red goal because of its action history despite the fact it is now reducing its distance to both goals. The intention estimation algorithm has to take action history into account when estimating the worker's intention.}
\label{fig:history}
\end{figure}

We use the distance vector \textbf{d} and the distance matrix \textbf{D} to validate worker's actions, and we  introduce the motion validation vector \textbf{v}, which is computed as follows:
\begin{equation}
\textbf{v} = \frac{\underset{1\leq i \leq n}\max{\textbf{D}_{ij}}-\textbf{d}}{\underset{1\leq i \leq n}\max{\textbf{D}_{ij}}-\underset{1\leq i \leq n}\min{\textbf{D}_{ij}}}.
\label{eq:validation}
\end{equation}
If the worker is moving towards the goal, the corresponding value of \textbf{v} will be close to unity, and if it is moving away from that goal, the corresponding value will gravitate to zero.
One could argue that  \textbf{v} may be interpreted as the estimate of worker intention, because it expresses a measure of approaching the goal.
However, elements of \textbf{v} are very sensitive to sensor noise and they would need to be filtered if one wanted to draw inferences about worker intentions directly from them.
Additionally, even though observing the current value of \textbf{v} does indeed carry crucial information for estimating worker intention, it is not sufficient since it lacks history of past values of \textbf{v}.
Consider the following simple example depicted in Fig.~\ref{fig:history}, where a warehouse worker moved past the goal labeled by red square, and further advanced towards the goal labeled with the green square.
As soon as the worker turned right, the value of \textbf{v} related to the green, as well as red, goal started to grow, since the distance between the worker and that goal started to drop.
However, since the worker previously failed to turn towards the red goal, its intention estimation for that goal should have remained low.
This example also demonstrates why there is a need to measure distance of traversable path as the approaching measure, instead of simply having, e.g., the Euclidean distance as elements of vector \textbf{v}.
Before further calculations we apply discrete first-order low-pass filter on \textbf{v} in order to reduce the noise influence.

\subsection{HMM based intention estimation}

\begin{figure}[!t]
\centering
\includegraphics[width=.95\linewidth]{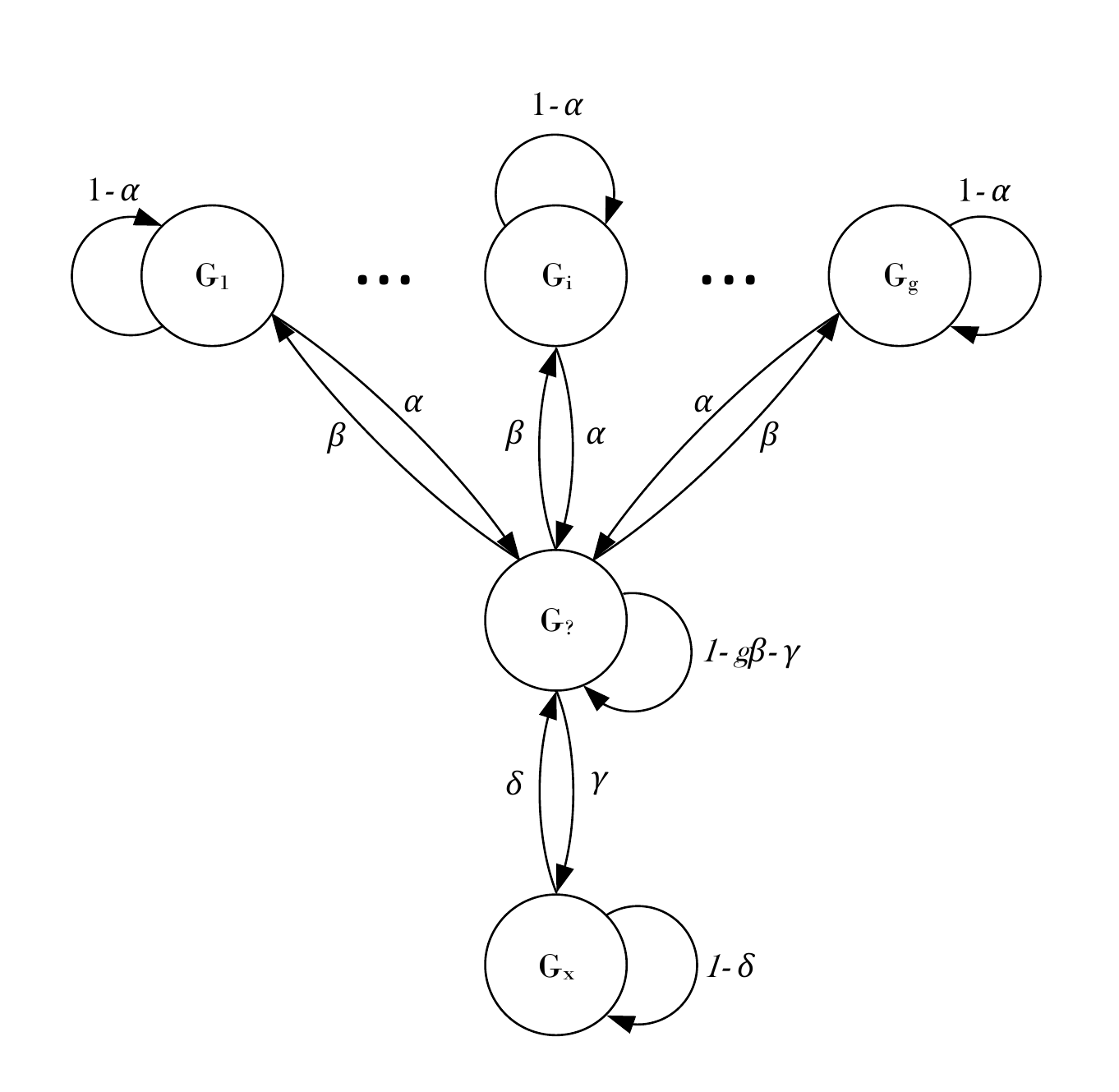}
\caption{HMM architecture used for human intention recognition. Worker's change of mind tendency is captured by the parameter $\alpha$ and the parameter couple $\beta$ and $\gamma$ set the threshold for estimating intention for each goal location. Increasing $\beta$ leads to quicker inference of worker's intentions and increasing $\gamma$ speeds up the decision making process. Parameter $\delta$ captures model's reluctance to return to estimating the other goal probabilities once it estimated that the worker is irrational.}
\label{fig:HMM_states}
\end{figure}

Given that we know how to calculate the motion validation vector \textbf{v}, we now introduce the model for solving the worker's intention estimation problem.
While worker's actions, manifested as moving and turning, are fully observable, they depend on the worker's inner states (desires), which cannot be observed and need to be estimated \cite{Bandyopadhyay2013}.
We propose a framework based on hidden Markov model for solving the worker's intention estimation problem.
HMMs are especially known for their application in temporal pattern recognition such as speech, handwriting, gesture recognition \cite{Rabiner1989a} and force analysis \cite{Chen1997}.
They are an MDP extension including the case where the observation (worker's action) is a probabilistic function of the hidden state (worker's desires) which cannot be directly observed.
We propose a model with $g+2$ hidden states shown in Fig.~\ref{fig:HMM_states} and listed in Table~\ref{tbl:HMM_states}.

\begin{table}[!t]
\caption{HMM framework components}
\label{tbl:HMM_states}
\centering
\begin{tabular}{lll} \toprule
Symbol & Name & Description\\ \midrule
$G_i$ & Goal $i$ & Worker wants to go to Goal $i$ \\
$G_?$ & Unknown goal & Worker's intentions are not certain \\
$G_x$ & Irrational worker & Worker behaves irrationally \\
\end{tabular}
\end{table}

Hidden states $G_i$ describe worker's intention of going to $i$-th goal, $G_?$ indicates that the worker prefers multiple goals and the model cannot decide on the exact desire with enough certainty.
On the other hand, hidden state $G_x$ indicates that the worker is moving away from all the goal locations.
This hidden state models the case of the worker being irrational or worker desiring a goal we have not yet specified.
The proposed model cannot distinguish between these two cases.
Introduced hidden states enable the human intention recognition model to elegantly save the intention estimation history as probabilities $P(G_i)$.
The first building block in this HMM architecture is the transition matrix $\textbf{T}^{g+2 \times g+2}$:
\begingroup
\renewcommand*{\arraycolsep}{5pt}
\begin{equation}
\textbf{T}=\begin{bmatrix}
1-\alpha& 0 & \dots & \alpha & 0 \\
0 & 1-\alpha &\dots & \alpha & 0 \\
\vdots & & \ddots & & \vdots \\
\beta&\beta& \dots & 1-g\beta-\gamma & \gamma \\
0 & 0 & \dots & \delta & 1-\delta \\
\end{bmatrix},
\label{eq:transition}
\end{equation}
\endgroup
where the architecture and description of the matrix parameters can be seen in Fig.~\ref{fig:HMM_states}. We have obtained parameters experimentally as follows: $\alpha=0.5$, $\beta=0.1$, $\gamma=0.05$, $\delta=0.1$.

We use the calculated motion validation vector \textbf{v} to generate the HMM's emission matrix $\textbf{B}$.
Every time the worker moves or turns significantly, we estimate the  worker intention using the Viterbi algorithm \cite{Forney1973}, which is often used for solving HMM human intention recognition models \cite{Zhu2008}.
The inputs of the Viterbi algorithm are the hidden states set $S=\{G_1, ... G_g, G_?, G_x\} $, hidden state transition matrix $\textbf{T}$, initial state $\Pi$, sequence of observations $\textbf{O}$, and the emission matrix $\textbf{B}$.
The HMM framework generally assumes a discrete set of observations, but since our observation is the validation vector $\textbf{v}$ with continuous element values, we have decided to modify the input to the Viterbi algorithm by introducing an expandable emission matrix $\textbf{B}$.
While the classic HMM emission matrix $\textbf{B}^{n \times g}$ links hidden states with discrete observations via fixed conditional probability values, elements of the introduced expandable emission matrix $\textbf{B}^{k \times g}$, where $k$ is the recorded number of observations, are functions of the observation value.
By using this modification of the emission matrix, we additionally simplify the Viterbi algorithm because there is no set of discrete observations it has to iterate through.
Once a new validation vector $v$ is calculated, the emission matrix is expanded with the row $\textbf{B}'$, where the element $\textbf{B}'_{i}$ stores the probability of observing \textbf{v} from hidden state $G_i$.
We also calculate the average of the last $m$ vectors \textbf{v} and the maximum average value $\phi$ is selected.
It is used as an indicator if the worker is behaving irrationally, i.e., is not moving towards any predefined goal.
The value of the hyperparameter $m$ decides how much evidence we want to collect before we allow the algorithm to declare the worker irrational.
If the worker has been moving towards at least one goal in the last $m$ iterations ($\phi>0.5$), we calculate $\textbf{B}'$ as:
\begin{equation}
B'=\zeta \cdot
\begin{bmatrix} \tanh(\textbf{v}) & \tanh(1-\Delta) & 0 \end{bmatrix}
\label{eq:Bmatrix},
\end{equation}
and otherwise as:
\begin{equation}
B'=\zeta \cdot
\begin{bmatrix} \boldsymbol{0}_{1\times g} & \tanh(0.1) & \tanh(1-\phi) \end{bmatrix},
\label{eq:Bmatrix2}
\end{equation}
where $\zeta$ is a normalizing constant and we calculate $\Delta$ as difference of the largest and second largest element of \textbf{v}.
Using such way of calculating $\Delta$ enables us to simply encode that, in order for our model to decide in favor for any goal location, it has to significantly stand out from other goals.
Humans often infer intentions of others by observing their actions \cite{Baker2014a}, which are generally not optimal with respect to their goals.
Nevertheless, we argue that the worker will globally move towards the goal it desires the most, but may locally take suboptimal actions such as looking around or swerving laterally.
In order to encode such behavior, we use the hyperbolic tangent function in \eqref{eq:Bmatrix} and \eqref{eq:Bmatrix2} to reduce the difference between the actions that indicate movement towards the goal whilst penalizing other actions approximately equal as linear function would.
Finally, we set initial probabilities of worker's intentions as:
\begin{equation}
\Pi= \begin{bmatrix}
0 & \dots & 0 & 1 & 0
\end{bmatrix},
\label{eq:initial_desires}
\end{equation}
indicating that the initial state is $G_?$ and the model does not know which goal the worker desires the most.
The Viterbi algorithm outputs the most probable hidden state sequence and the probabilities $P(G_i)$ of each hidden state in each step.
These probabilities are the worker's intention estimates.

Worker tasks are not always predefined in the beginning and can appear or cease during worker's stay on the shop floor.
We have taken such events into consideration and made it possible to add or remove goals during the experiment.
If the goal has to be removed, e.g., because other worker took that job over or the task was canceled, we simply discard that goal from our calculations and add it to the unknown goal intention estimation.
If the goal has to be added, we do this by setting its intention estimation to $\max({\min({P(G_i}), 0.1)}$ with $0<i<g$ and by expanding the \textbf{I} and \textbf{T} matrices.
We would like to emphasize that it is not possible to add an arbitrary number of goals, because of the fixed values of parameters in \textbf{T} matrix.
The maximum number of goals this model can handle is limited because elements of transition matrix \textbf{T} must be greater than 0.
The only value of matrix \textbf{T} that depends on goal number is probability of staying in $G_?$ state and equals to $1-g\beta-\gamma$.
We calculate the maximum number of goals by applying positivity condition to that expression:
\begin{equation}
g_{max}= \floor[\Big]{ \frac{1-\gamma}{\beta} },
\label{eq:max_goals}
\end{equation}
which for our current value of parameters amounts to 9 goals.
However, after detailed testing, we have concluded that proposed model does not perform well if there are more than 5 goals.
The main difficulty here is that if there are more potential goals, worker's actions are not discriminatory enough and there is not enough evidence that the worker desires one goal more than the other goals.
Because of that, the model estimates $G_?$ as the most probable state throughout the experiment.
While we argue that this still is accurate worker's intention estimation, it is hardly useful to supervisory system that is taking our algorithm's estimations as input.

\section{Experimental Results}\label{sec:experimental_results}

\begin{figure*}[!t]
	\vspace{-0.35cm}
	\centering
	\includegraphics[width=0.9\textwidth]{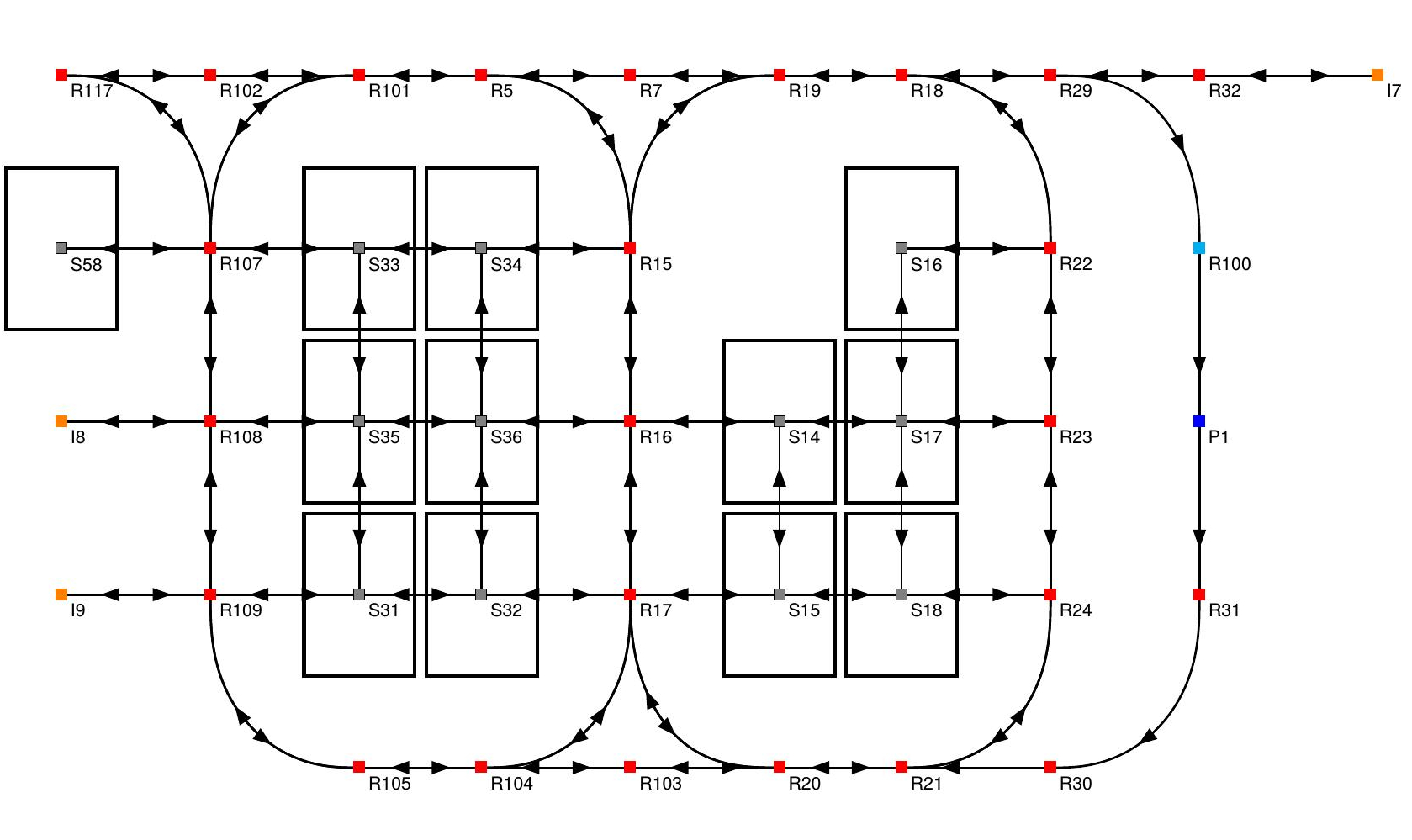}
	\caption{The layout of the laboratory warehouse used for AR-based experiments. Mobile robots move over ground nodes labeled with letters R (traversable path for warehouse workers) and S (under the racks which robots can pick up and move). Nodes P and R100 are used as picking station and queue node but are treated the same as other R nodes in our experiment setup.  }
	\label{laboratory_layout}
\end{figure*}

In this section we demonstrate proposed human intention estimation algorithm performance and discuss the results.
We have conducted experiments in both an industrial setup and in larger virtual reality rendered warehouse.
Before we analyze the results, we discuss the method for evaluating the proposed model.

To the best of our knowledge, a recognized criterion or method for evaluating human intention estimation models does not exist.
Models such as those proposed in \citep{Baker2014a} rely on people's judgments to evaluate results, while models which use learning methods have a well defined start and end points of the experiments and the ground truth information is readily available or evident from the experiment's end point.
While worker's intention at the end of the experiment is unambiguous, it is unclear how to empirically determine intentions \emph{during} the experiment because of the possibility of worker's change of mind.
Also, we obtained the parameters given in \eqref{eq:transition} experimentally after thorough testing, and they were selected in a way to produce consistent and semantically interpretable results on different datasets.
We do not claim that those parameters are in any way optimal, but we assert that applying some of the well-known algorithms for learning HMMs, such as expectation maximization, is not feasible in our case, since an unbiased labeled dataset is unavailable.
Furthermore, we deterred from using people's judgments as ground truth for the same reason.
Since we want estimates of the proposed human intention estimation algorithm to be useful to a supervisory system, we insisted that if the worker is moving towards more than one goal, the intention for those goals should be higher than for any other goals; but, it should be lower than estimation for $G_?$.
If the worker moves to a single goal, intention for that goal should be the highest, while if the worker does not move towards any goal for a predefined amount of time, the model should declare it as irrational.
We have chosen that time to be 1 second.
Using such interpretation of intentions can be encoded in a mobile robot fleet management system, which can then, e.g., reroute mobile robots away from the goal worker is moving to.
We assert that introduction of $G_?$ state keeps the algorithm's response time low, but simultaneously  encodes measure of uncertainty of the intention estimation for goal locations.
\begin{figure*}[!h]
\centering
\subfloat[First-person view from the 2 Megapixel Hololens camera\label{AR_exp}]{\includegraphics[width=0.48\textwidth]{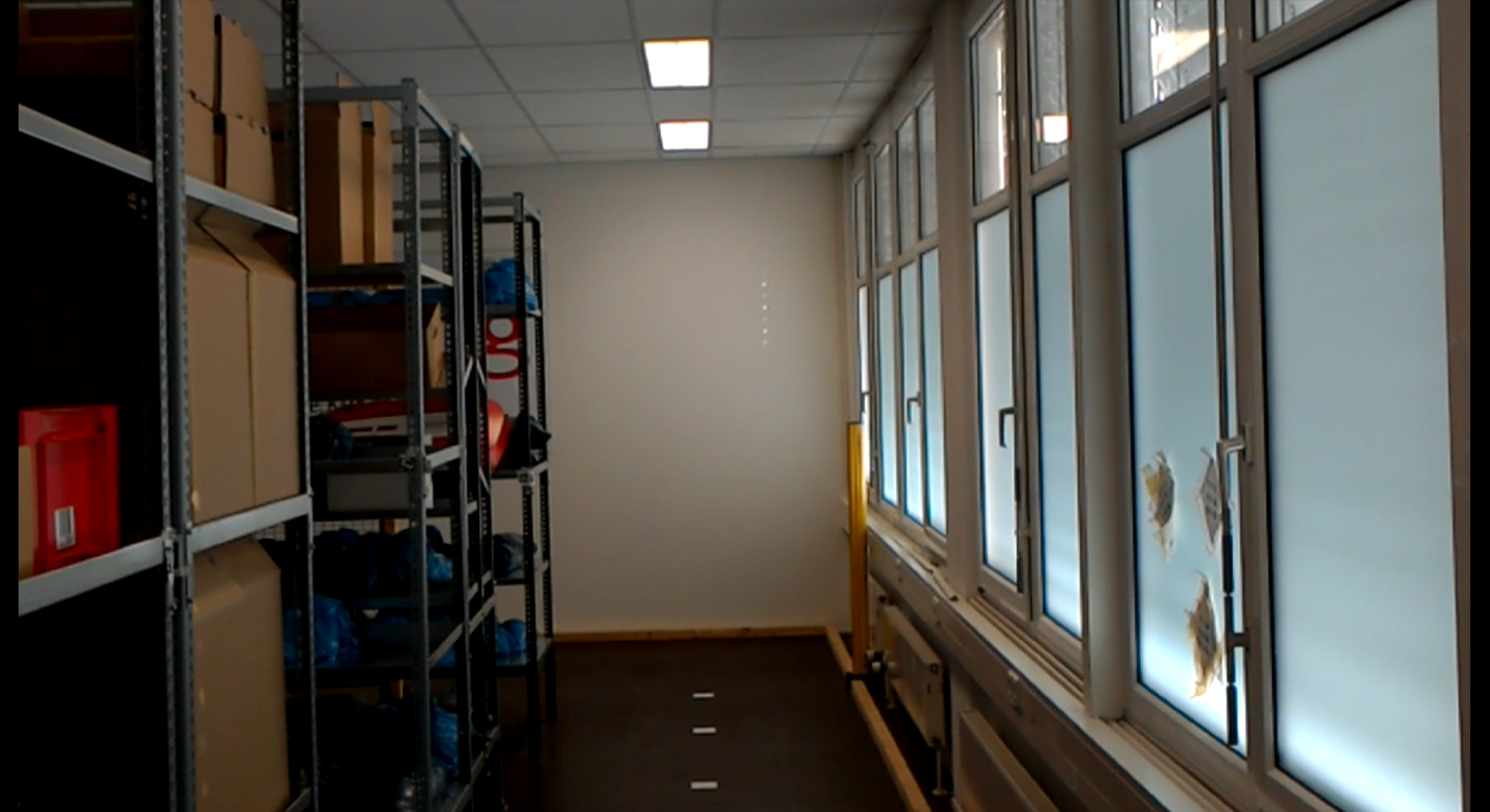}}
\qquad
\subfloat[First-person view from HTC Vive\label{VR_exp}]
{\includegraphics[width=0.48\textwidth]{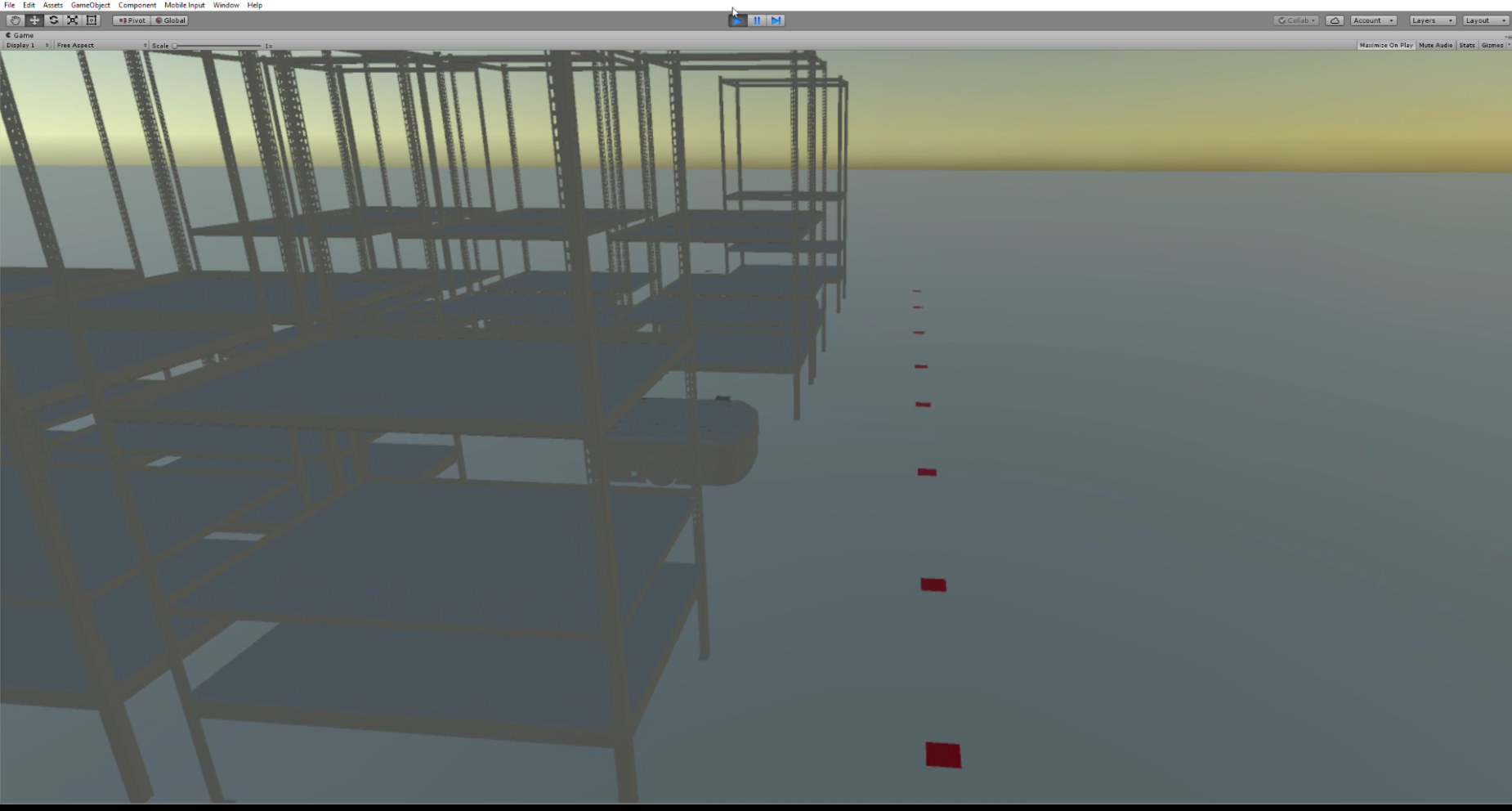}}
\caption{Comparison between the real-world experiment and the experiments in VR. Both experiments use exactly the same setup.}
	\vspace{-0.35cm}
\label{AR_VR}
\end{figure*}
\subsection{Augmented reality experiments}
\begin{figure}[htb]
	\centering
	\includegraphics[width=0.47\textwidth]{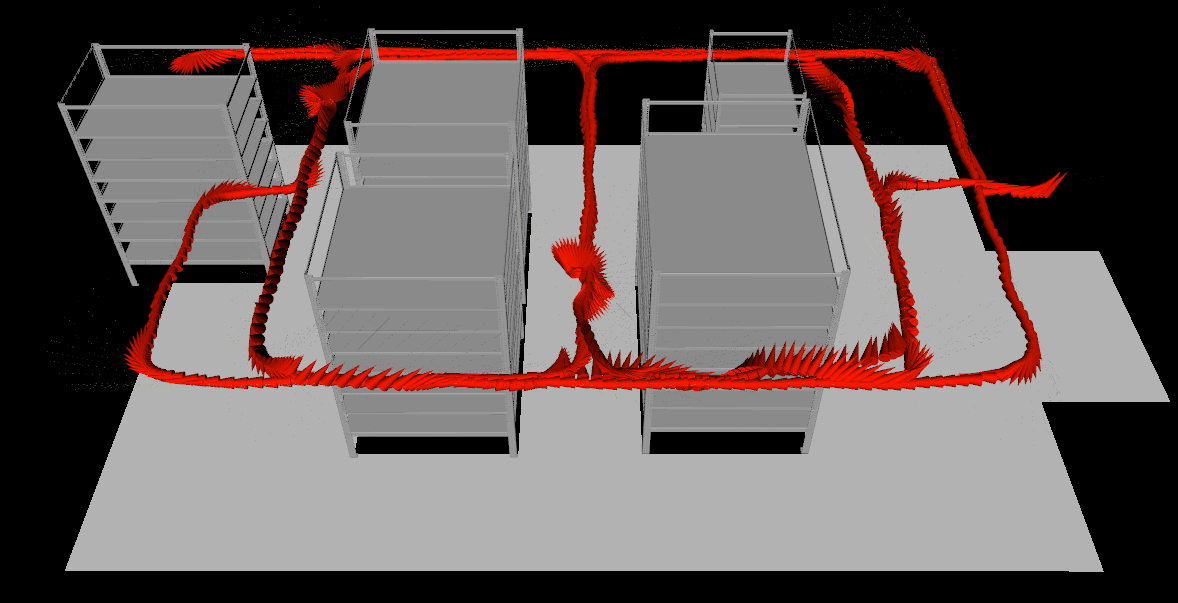}
	\caption{Conducted experiment showcasing the precision of the Hololens' localization. We have created a warehouse model in \textit{rviz} - Robot Operating System's 3D visualization tool, which will be further used to demonstrate proposed algorithm's results.}
	\vspace{-0.35cm}

	\label{localization}
\end{figure}
The augmented reality experiments were conducted in the laboratory warehouse of Swisslog and consists of twelve racks and two robots.
The layout of the warehouse is shown in Fig.~\ref{laboratory_layout}.
Although of smaller scale than commercial warehouses, it nevertheless enables conducting experiments in a realistic environment.
Since our intention estimation algorithm assumes that position and orientation of the worker are known, we used the Hololens' SLAM algorithm to localize the person while walking inside the warehouse.
Since no external tracker was used, we fist wanted to verify the localization accuracy of the Hololens inside a warehouse-like environment.
The experiment was conducted by a person first starting from a predefined ground node (R32 in Fig.~\ref{laboratory_layout}, all the ground nodes are manually marked and unique) and then walked from ground node to ground node.
In this experiment, all AR interactions were disabled and the person navigated the warehouse autonomously and the only exception was a holographic sphere positioned at the Hololens' location every 100 frames.
This allowed us to conduct a qualitative analysis of the localization shown in Fig.~\ref{localization}.
The results showed that the localization was indeed robust with only a few centimeter deviation, at most, from the straight line paths between nodes.
A first-person view from the Microsoft Hololens is shown in Fig.~\ref{AR_exp}.

\begin{figure*}[!t]
\subfloat[Initial position.]{\includegraphics[width= .33\textwidth]{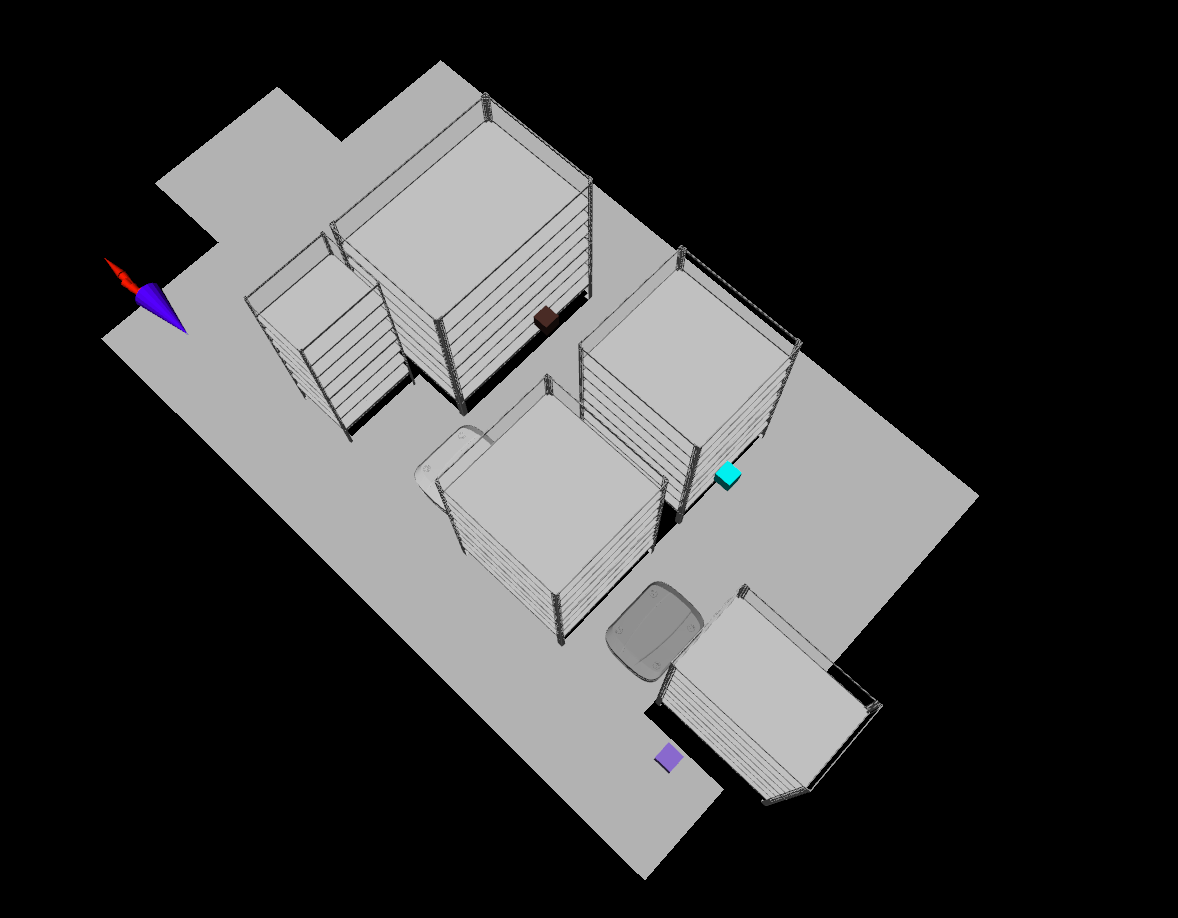}} \hspace*{.0015\textwidth}
\subfloat[The worker moves towards the purple goal (R117) but the mobile robot obstructs the intended path towards it. Because the worker has not turned around immediately, the model warns that the worker is behaving irrationally.]{\includegraphics[width= .33\textwidth]{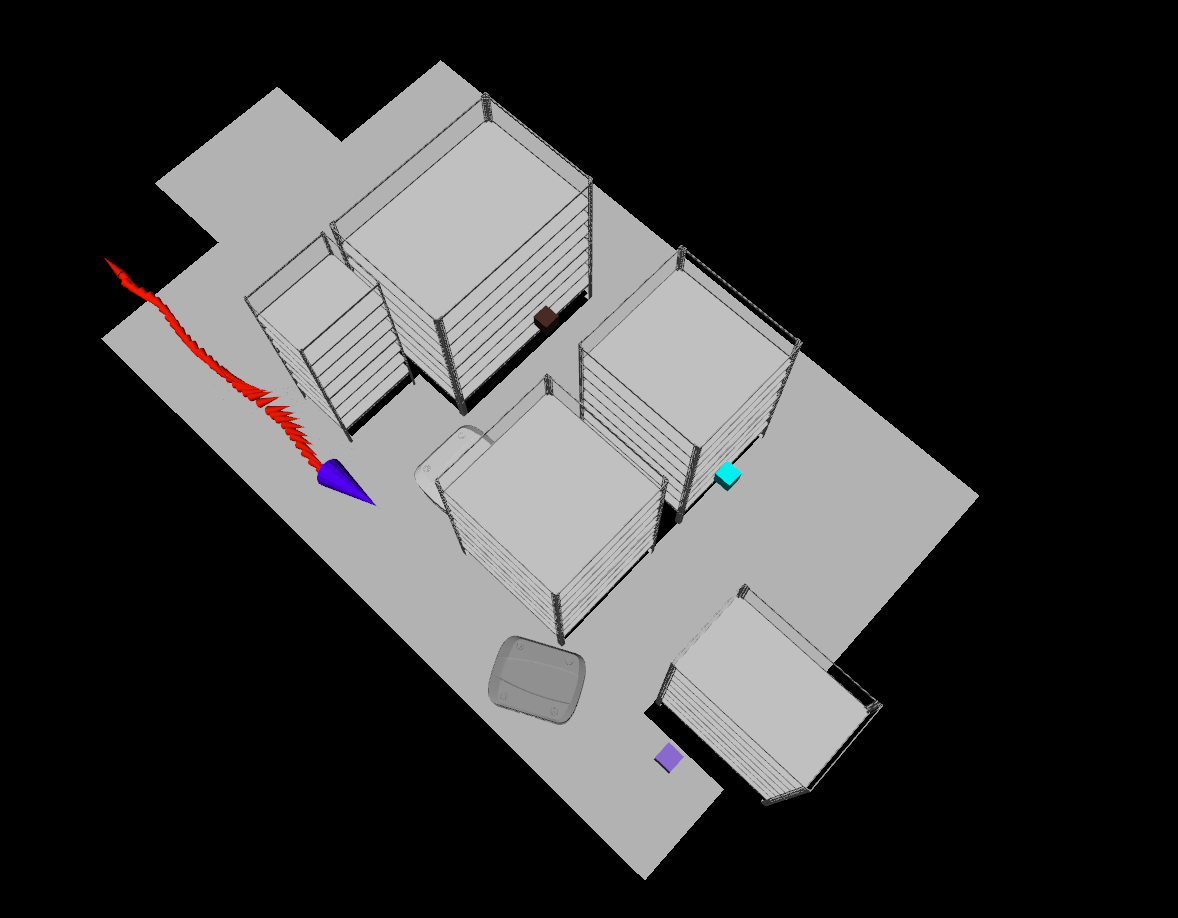}}\hspace*{.0015\textwidth}
\subfloat[The worker turns and follows the path consistent with going to all three goals.]{\includegraphics[width= .33\textwidth]{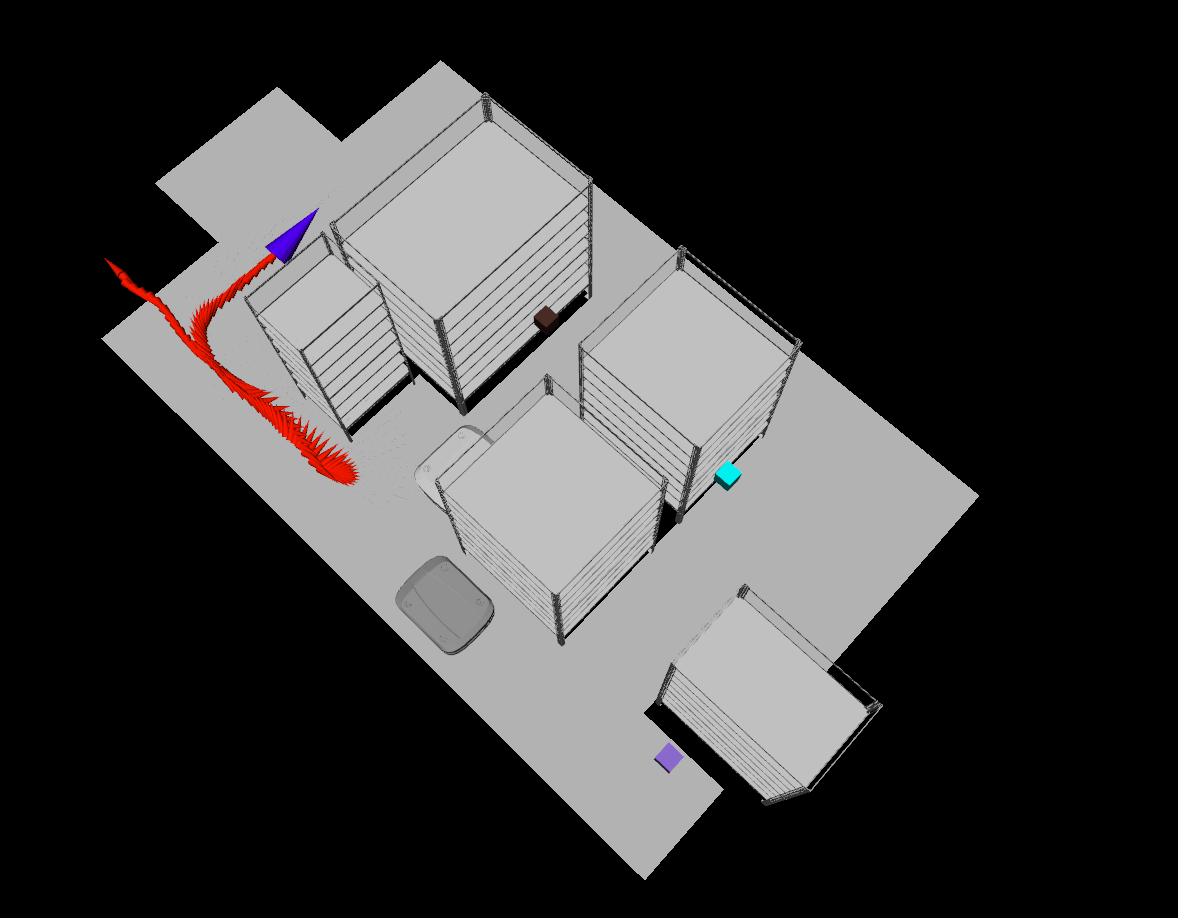}} 
\vspace{-0.35cm}
\subfloat[The worker is on a crossroad. If it turns towards the brown goal it is obvious that it is the goal wants to reach.]{\includegraphics[width= .33\textwidth]{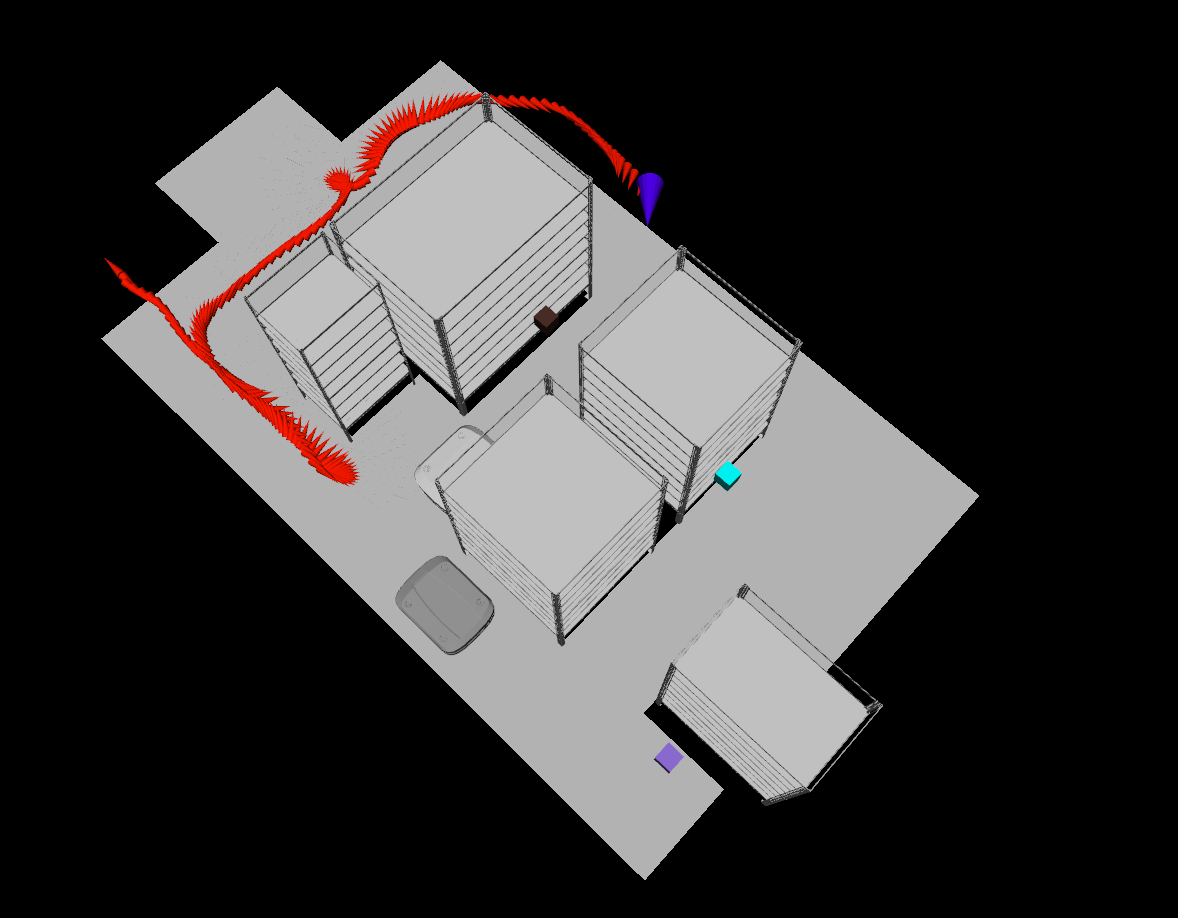}}\hspace*{.0015\textwidth}
\subfloat[However, the worker has decided not to advance towards the brown goal and to continue towards the cyan and purple goals instead. Because of that the intention estimation for the brown goal has steeply fallen.]{\includegraphics[width= .33\textwidth]{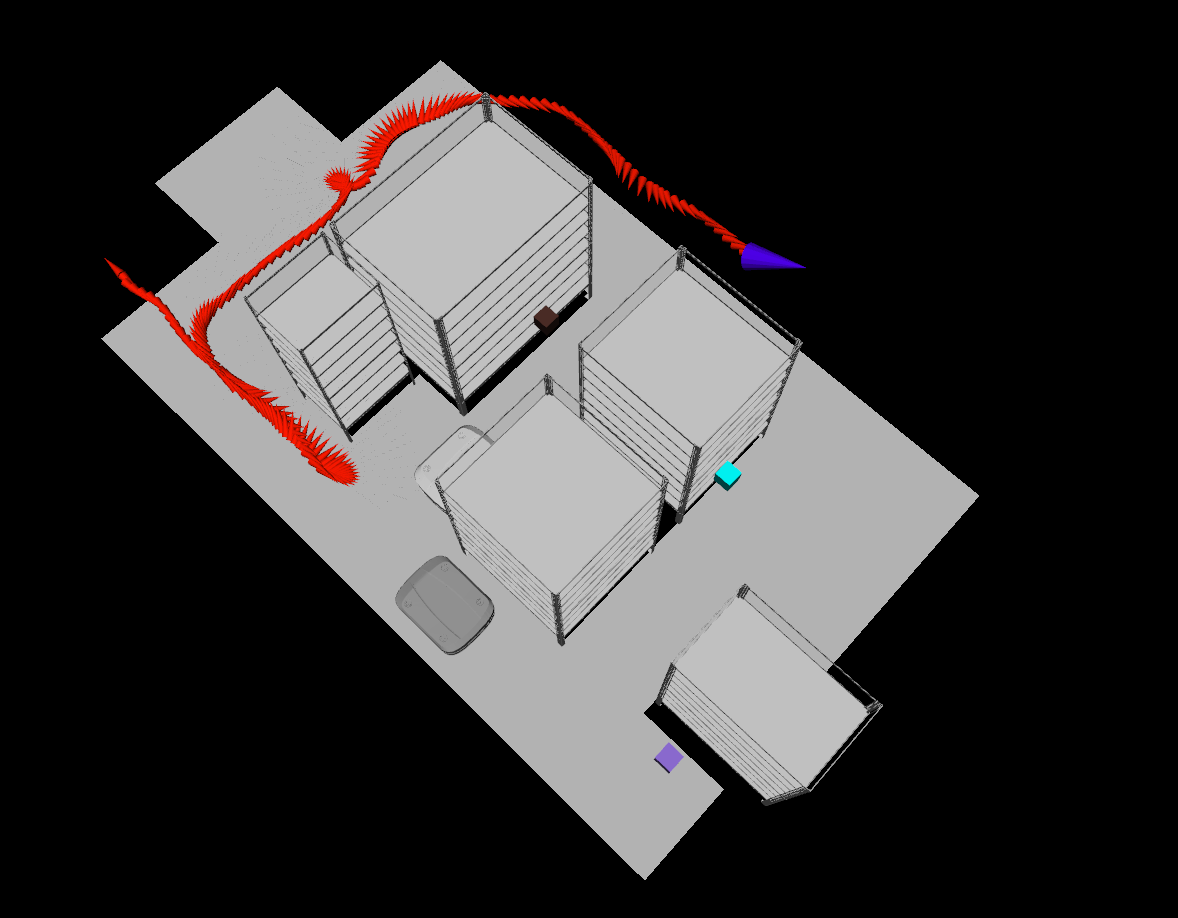}}\hspace*{.005\textwidth}
\subfloat[The worker stopped near the cyan goal and started turning around in place.]{\includegraphics[width= .33\textwidth]{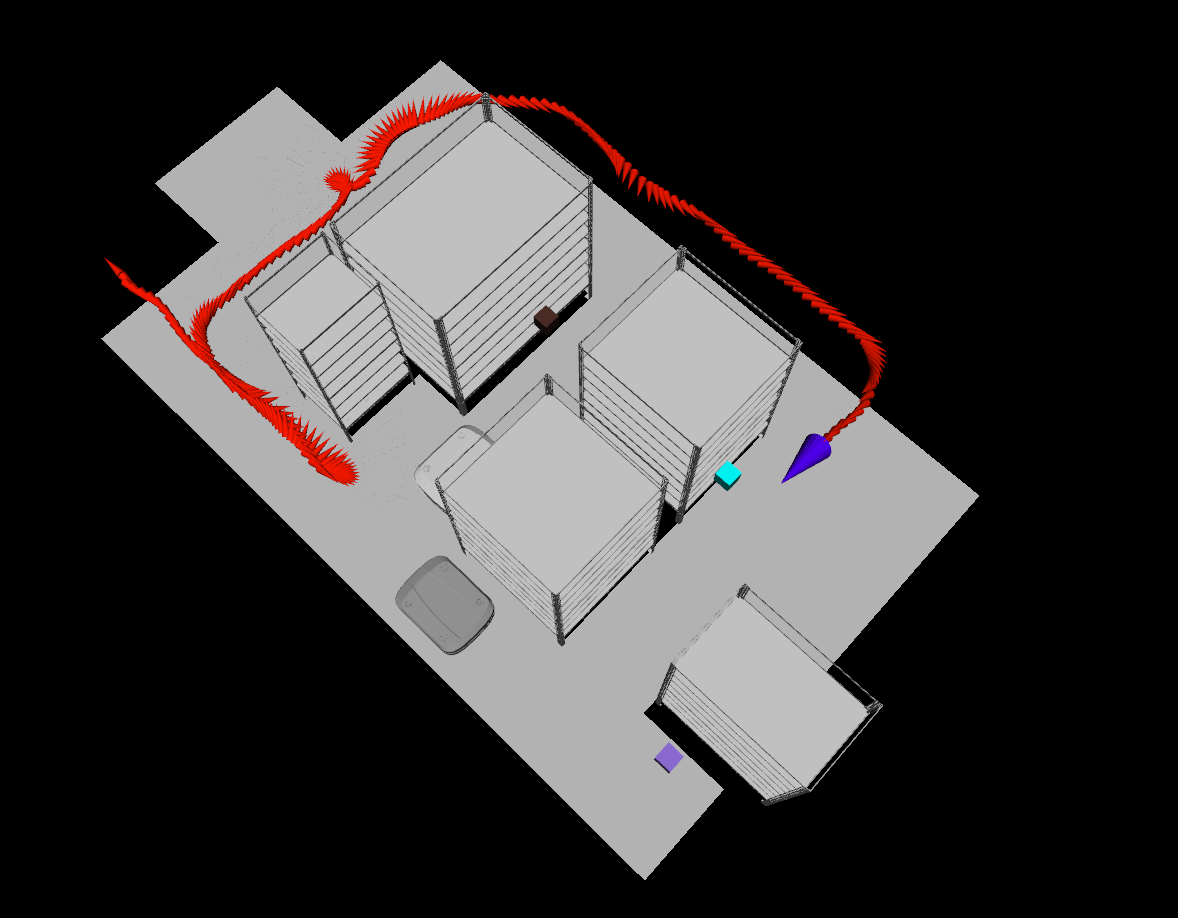}}
\vspace{-0.35cm}

\subfloat[The worker has taken its AR device off and is moving towards the brown goal.]{\includegraphics[width= .33\textwidth]{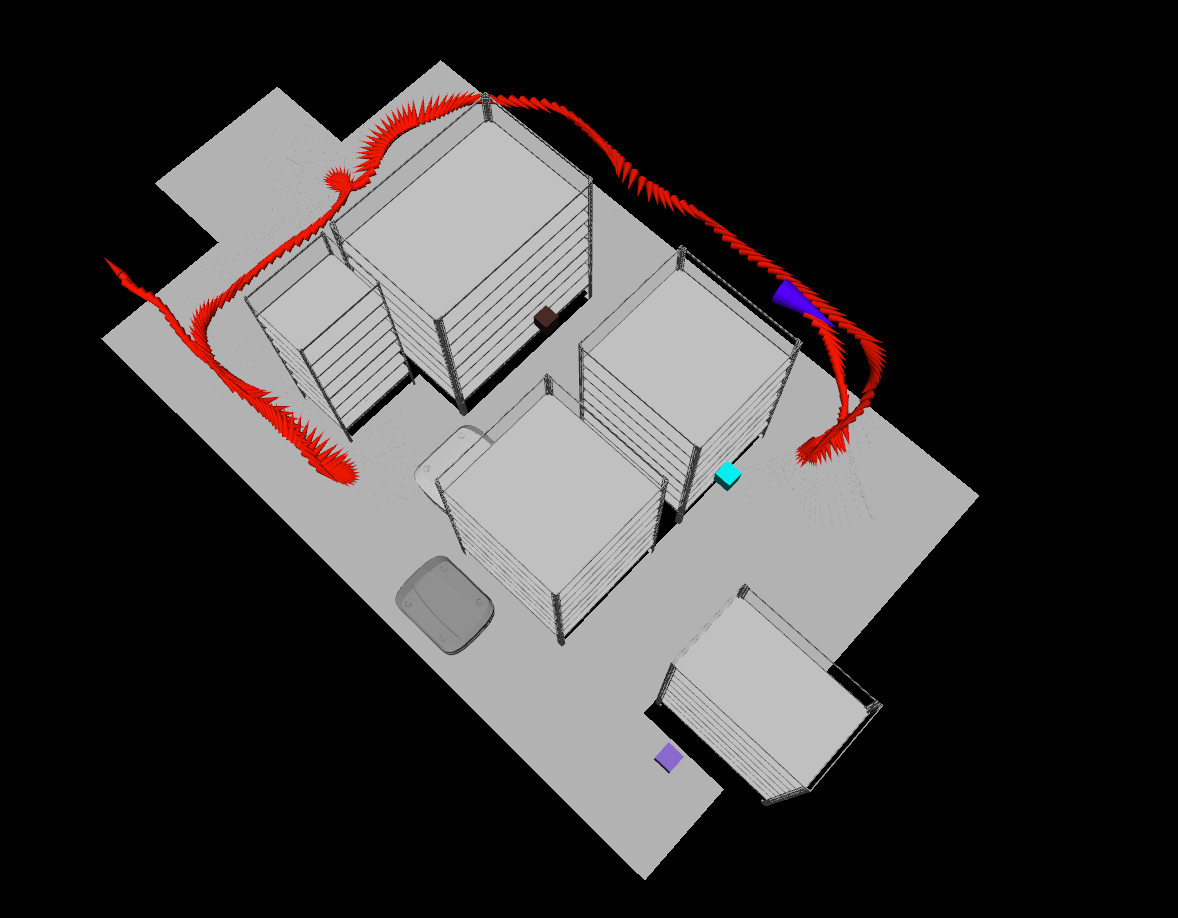}}\hspace*{.0015\textwidth}
\subfloat[Because the worker passed brown goal and is moving backwards, the model estimates its behavior to be irrational.]{\includegraphics[width= .33\textwidth]{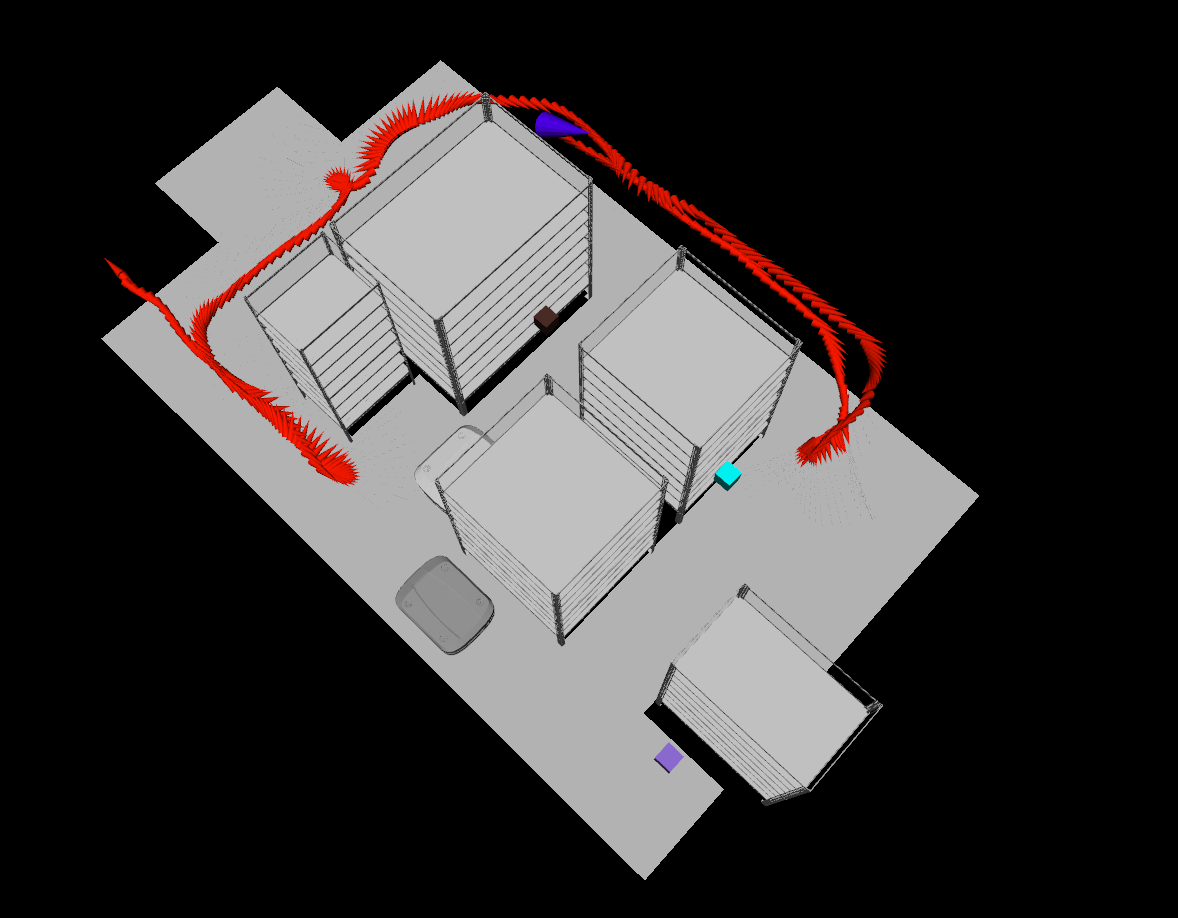}}\hspace*{.0015\textwidth}
\subfloat[End of the experiment.]{\includegraphics[width= .33\textwidth]{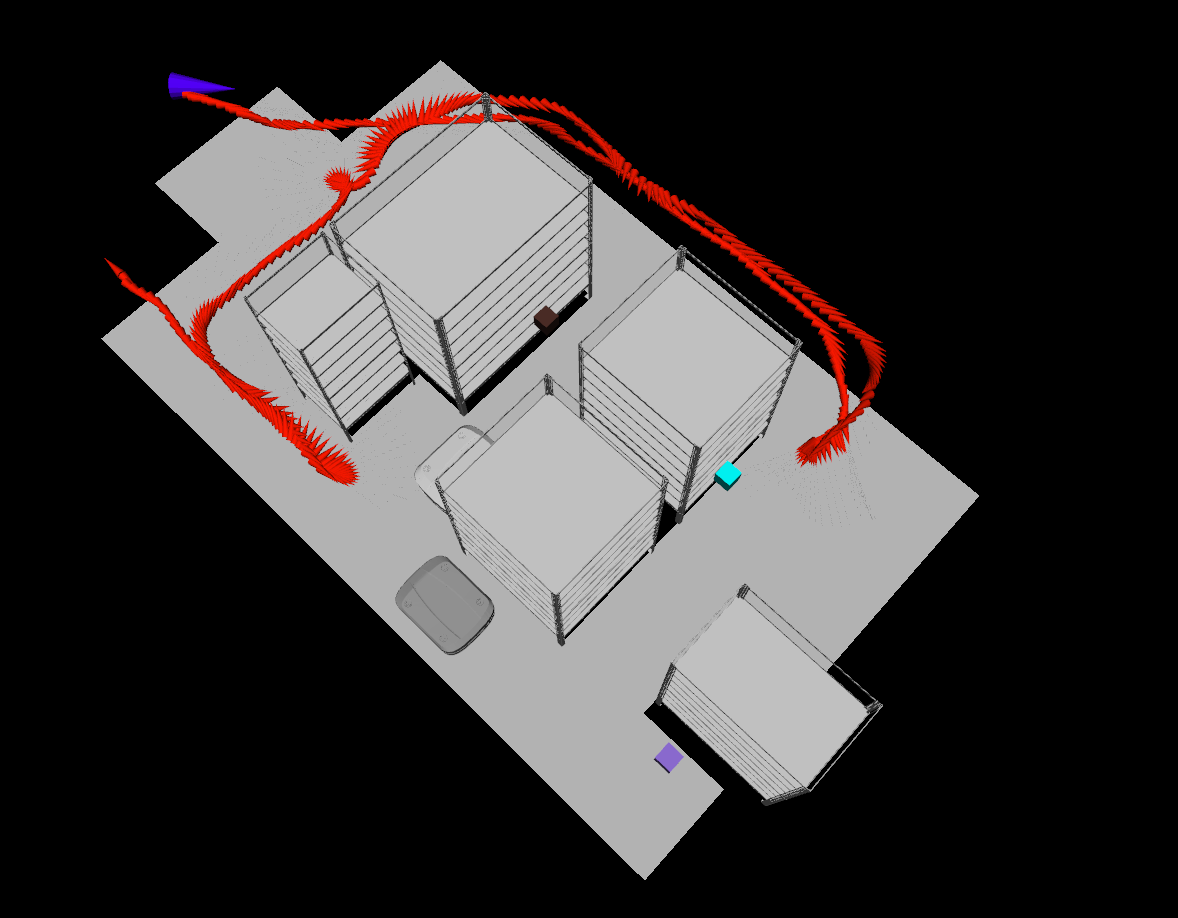}}
\vspace{-0.35cm}

\caption{Key moments of the industrial experiment setup in the laboratory warehouse visualized in \emph{rviz}. Goal nodes are labeled with cubes as follows: R117 purple, R17 brown and R109 cyan.}
\vspace{-0.35cm}

\label{fig:ettlingen_results}
\end{figure*}

\begin{figure*}[!t]
\centering
\subfloat[Algorithm's output in real world AR experiment.\label{probabilities_AR}]{
  \includegraphics[width=.9\columnwidth]{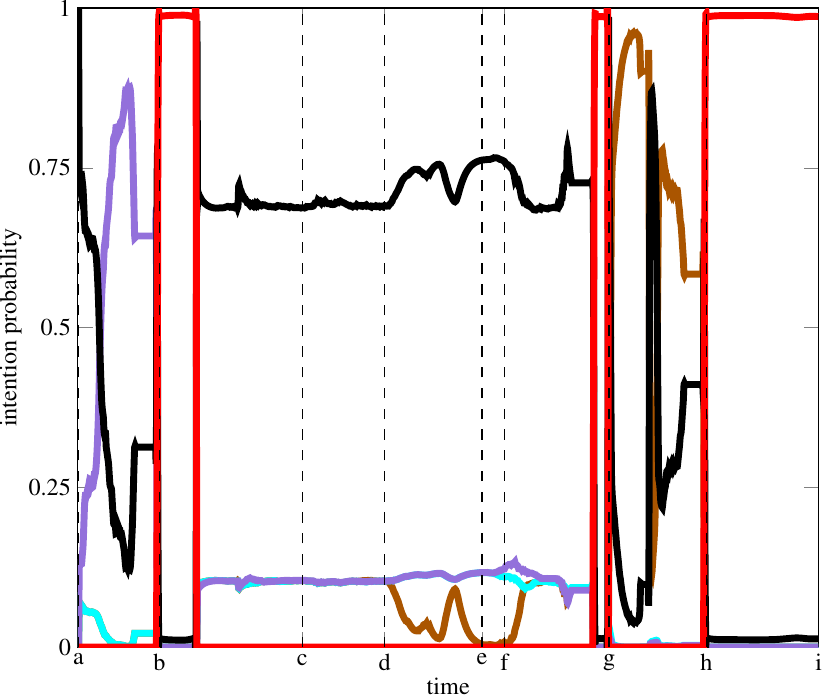}
  }
  \qquad
\subfloat[Algorithm's output in VR experiment.\label{probabilities_VR}]{
  \includegraphics[width=.9\columnwidth]{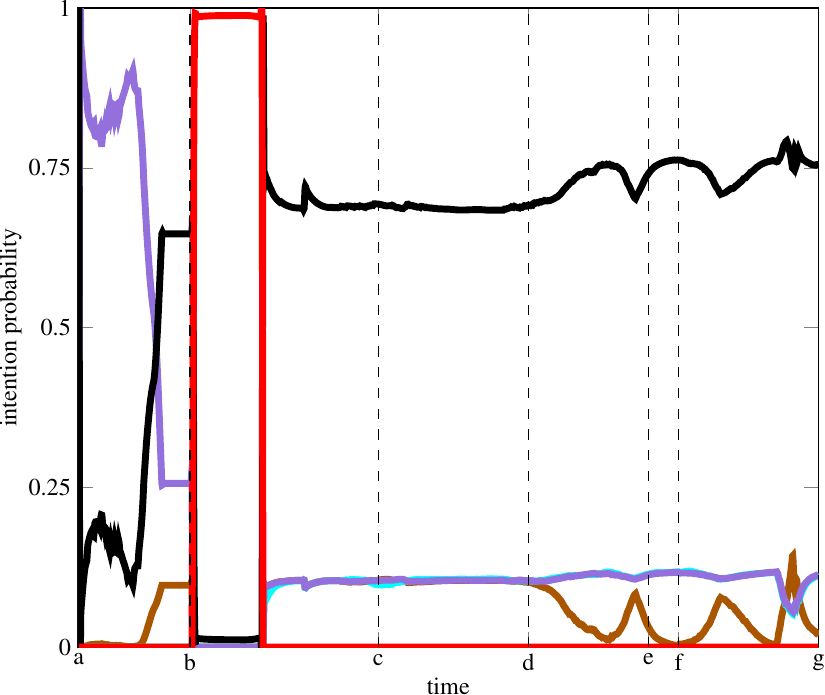}
  }
\caption{Comparison of the proposed algorithm's output between the real-world experiment and the experiment recreated in VR. Intention estimations for the three goal locations are labeled with respect to their color in Fig.~\ref{fig:ettlingen_results} (brown, cyan and purple), the unknown goal state is labeled black and the irrational worker state is labeled red. One can see that the results in this scenario are similar for both AR and VR technology.}
\vspace{-0.35cm}

\label{probabilities}
\end{figure*}

To test the proposed algorithm, two experiments were conducted (besides the localization experiment described previously).
In both experiments the starting point was node R32 and featured two robots; one stationary and positioned at R15, while the other was mobile.
In the first experiment shown in Fig.~\ref{fig:ettlingen_results} the worker has three potential goals located in front of racks near nodes on R117, R108 and R17.
The worker initially starts moving towards node R117, but the mobile robot moves forward from node R107 and turns right towards node R101, thus blocking the worker's initially intended path towards the R117.
The worker then turns around and follows the only remaining path towards R117 which is also the best path for the other two goals.
Once the worker reached R103 and continued to R104, the proposed algorithm detects that it has failed to turn towards the R17 goal and lowers its intention estimation value.
The worker further continues to goal R109 but turns around before reaching it.
After turning, the only goal the worker could be going to is R17, but the worker continues past it returning to node P. When worker passed the goal R17, the model recognized there are no goals it could be going to and declared the worker irrational.
The estimated intentions by the proposed algorithm can be seen in Fig.~\ref{probabilities_AR}.
The second experiment included a worker moving towards R117 and changing its mind to go to R17 simultaneously with mobile robot blocking the best path towards that goal and eventually returning to R117.
The main idea of the second experiment was to showcase algorithm's flexibility in scenarios where the worker changes its mind often.
Given that, we only show the second experiment in the accompanying video\footnote{https://youtu.be/SDD-v-pH0v4}.

\subsection{Virtual Reality Setup}\label{sec:vr_setup}

We built a virtual reality framework for rapid prototyping of applications for flexible robotized warehouses, a tool that might also evolve into a training framework in the future.
The main motivation behind a virtual framework, i.e., a virtual reality digital twin of a warehouse, is that  commercial automated warehouses are often unavailable for experiments, but they can be simulated, thus avoiding the warehouse downtime and financial losses, yet enabling testing in full scale to identify potential problems and obtain realistic user experience.
Furthermore, this approach also enables us to do tests with multiple users more freely and achieve the best possible interaction modalities.
Note that besides testing user or worker behavior inside the warehouse, the virtual setup also serves for testing augmented reality applications for warehouse workers wearing glasses such as the \emph{Microsoft Hololens}.
Concretely, the application was developed in \textit{Unity3D} and is used with the \textit{HTC Vive} headset.
An example of a user view within the virtual reality warehouse is shown in Fig.~\ref{VR_exp}.

The warehouses layout is planned using Swisslog proprietary network planner, from which an XML file is exported.
We have developed an XML file parser which together with an available CAD model builds a VR warehouse from scratch.
The fleet of robots is presently controlled by parsing a series of JSON messages, which are the output of a path planner \cite{Kulich}.

Currently two AR interaction scenarios have been implemented for prototyping augmented reality applications: path visualization for worker navigation and rack object picking assistance.
The virtual setup emulates the Microsoft Hololens: all of the \emph{holographic objects} are only visible through a narrow field of view (FoV), which according to specifications ranges from 30\degree{} to 35\degree{} horizontally and 17.5\degree{} to 17.82\degree{} vertically (we selected the lower bounds).
The interaction pointer uses raycasting to position itself with respect to objects.
Potentials of the proposed system are numerous.
For example, such a system allows us to test if a certain interaction modality fails or is not as informative because of the low FoV in a realistic environment.
It also allows us to inject localization errors to determine at which point various interaction modalities become unusable.
However, in the present paper the virtual framework is used primarily as a controlled simulated warehouse environment for conducting experiments for worker intention estimation.

\subsection{Virtual reality experiments}
Due to unavailability of a full-scale commercial warehouse for testing, we conducted larger-scale tests in VR, using the system described in Section~\ref{sec:vr_setup}.
The tracking method of the HTC Vive has an RMS error of 1.9\,mm, which offers very accurate tracking for a realistic VR experience for warehouse localization and worker behavior purposes.
We tracked the position of the user, as well as of each robot.
An example of a first-person view from the Hololens used in the laboratory warehouse and HTC Vive in a VR digital twin of the same warehouse can be seen in Fig.~\ref{AR_VR}.

\begin{figure}[htb]
	\centering
	\includegraphics[width=0.47\textwidth]{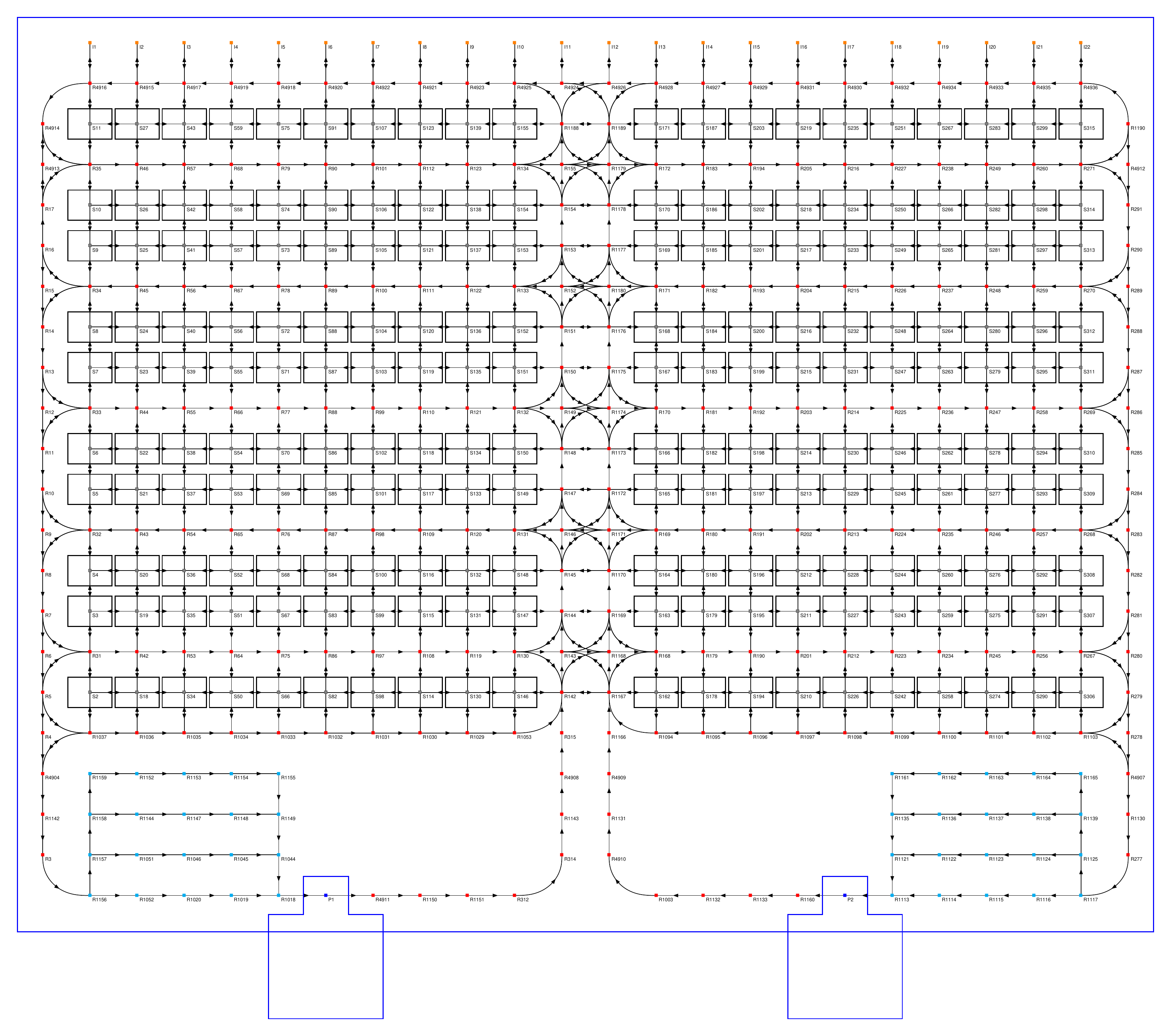}
	\caption{The larger warehouse layout used for VR experiments.}	\label{small_layout}
	\vspace{-0.35cm}

\end{figure}

We repeated the first AR experiment in the constructed VR environment to demonstrate applicability of VR for human intention estimation.
The only difference between scenarios was that we did not reproduce last part of the AR experiment when worker took off the AR device off to simulate a type of an irrational behavior.
The results are shown in Fig.~\ref{probabilities_VR}, where we can see that the biggest difference between real world and VR experiments is at the beginning of the experiment.
Those differences are caused by the fact that VR experiment starts with the worker going directly towards the goal, whereas at the beginning of the real word experiment the worker is still putting on the AR device.
Because they can be interpreted as semantically similar to AR results, we conclude that testing intention estimation in VR can produce credible results.

\begin{figure*}[!t]
\subfloat[Initial position.]{\includegraphics[width= .33\textwidth]{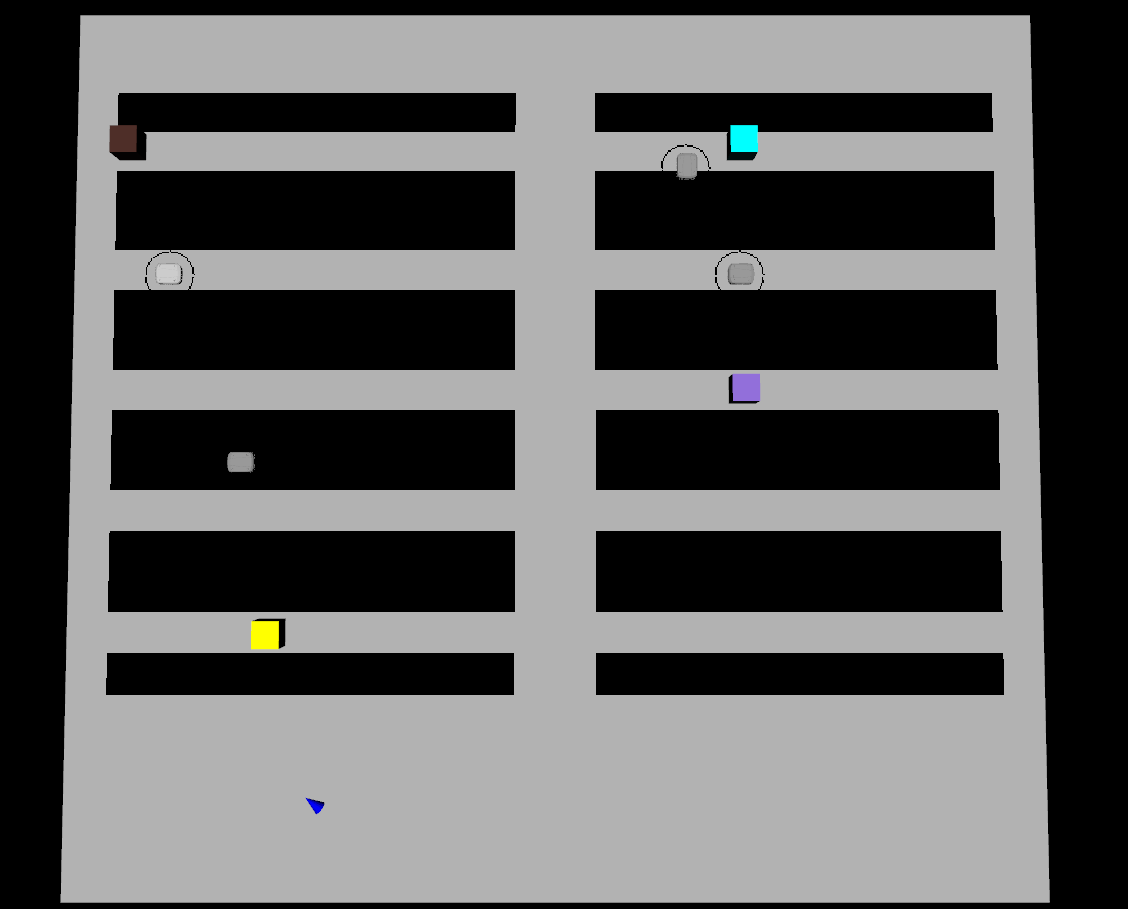}} \hspace*{.0015\textwidth}
\subfloat[The worker moves towards the yellow brown goals.]{\includegraphics[width= .33\textwidth]{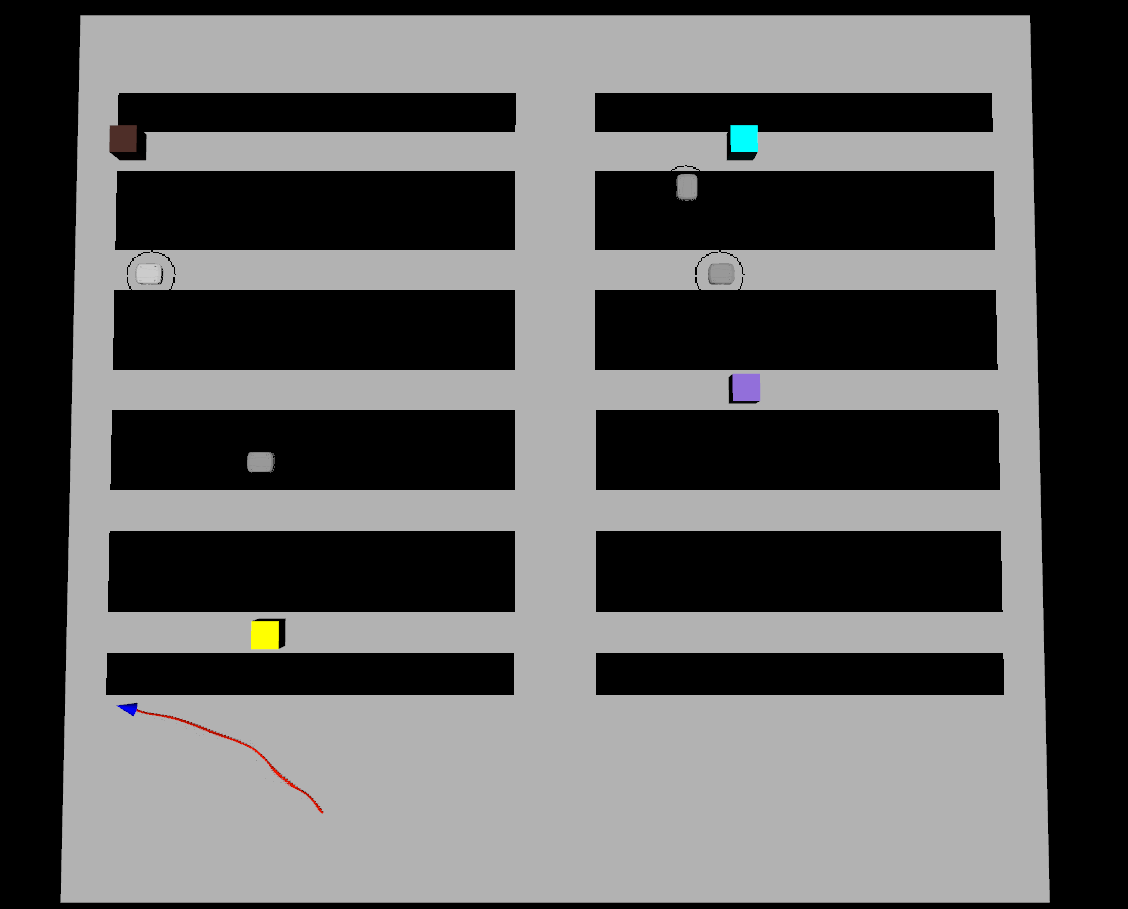}}\hspace*{.0015\textwidth}
\subfloat[The worker moves past the yellow goal.]{\includegraphics[width= .33\textwidth]{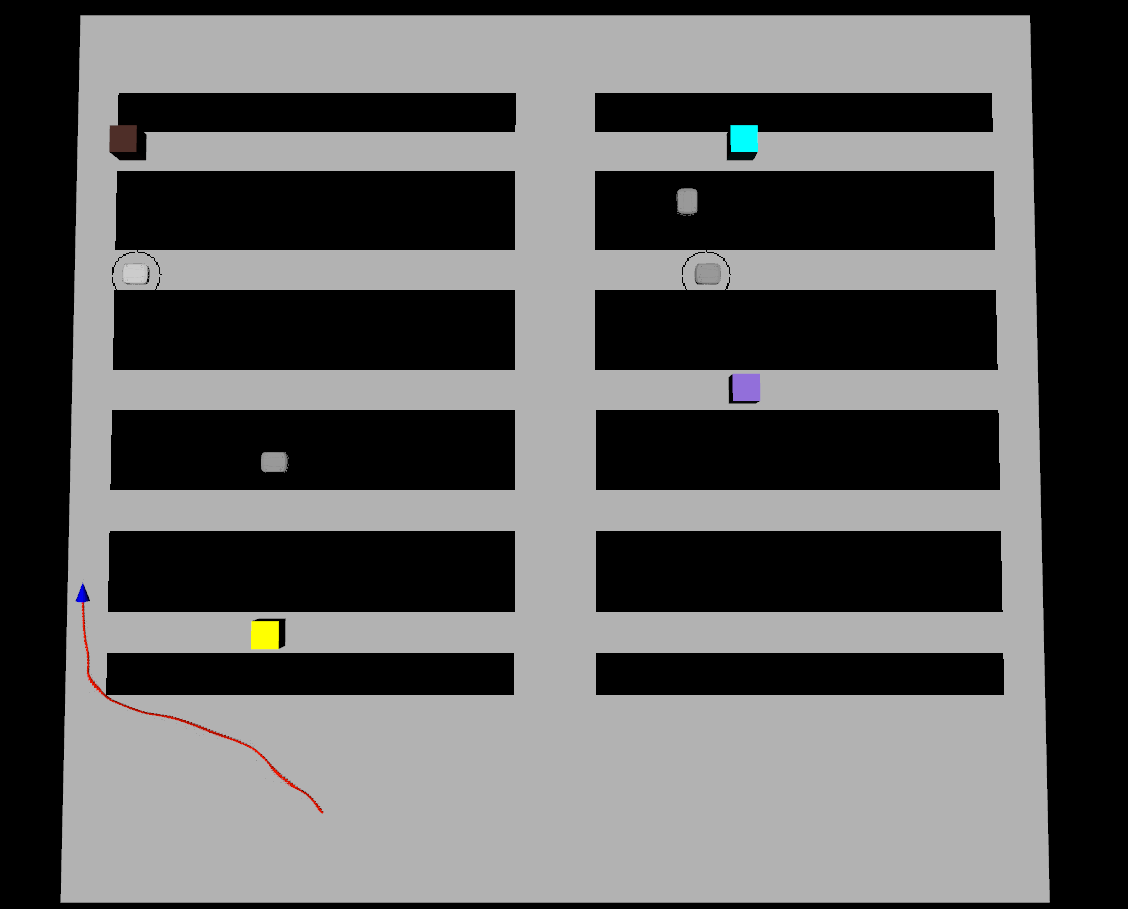}} 
\vspace{-0.35cm}

\subfloat[The brown goal is added in the second scenario.]{\includegraphics[width= .33\textwidth]{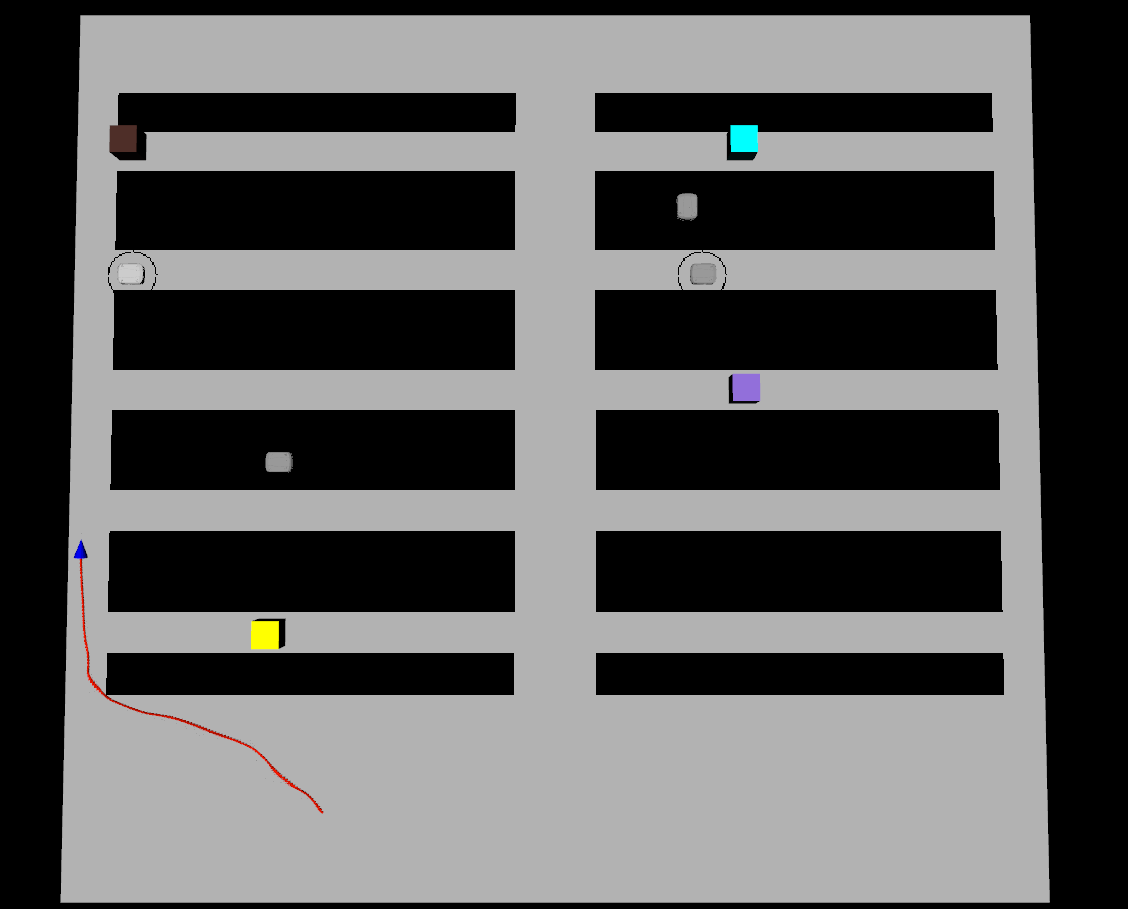}}\hspace*{.0015\textwidth}
\subfloat[A mobile robot blocks the worker's path towards the brown goal and the worker turns.]{\includegraphics[width= .33\textwidth]{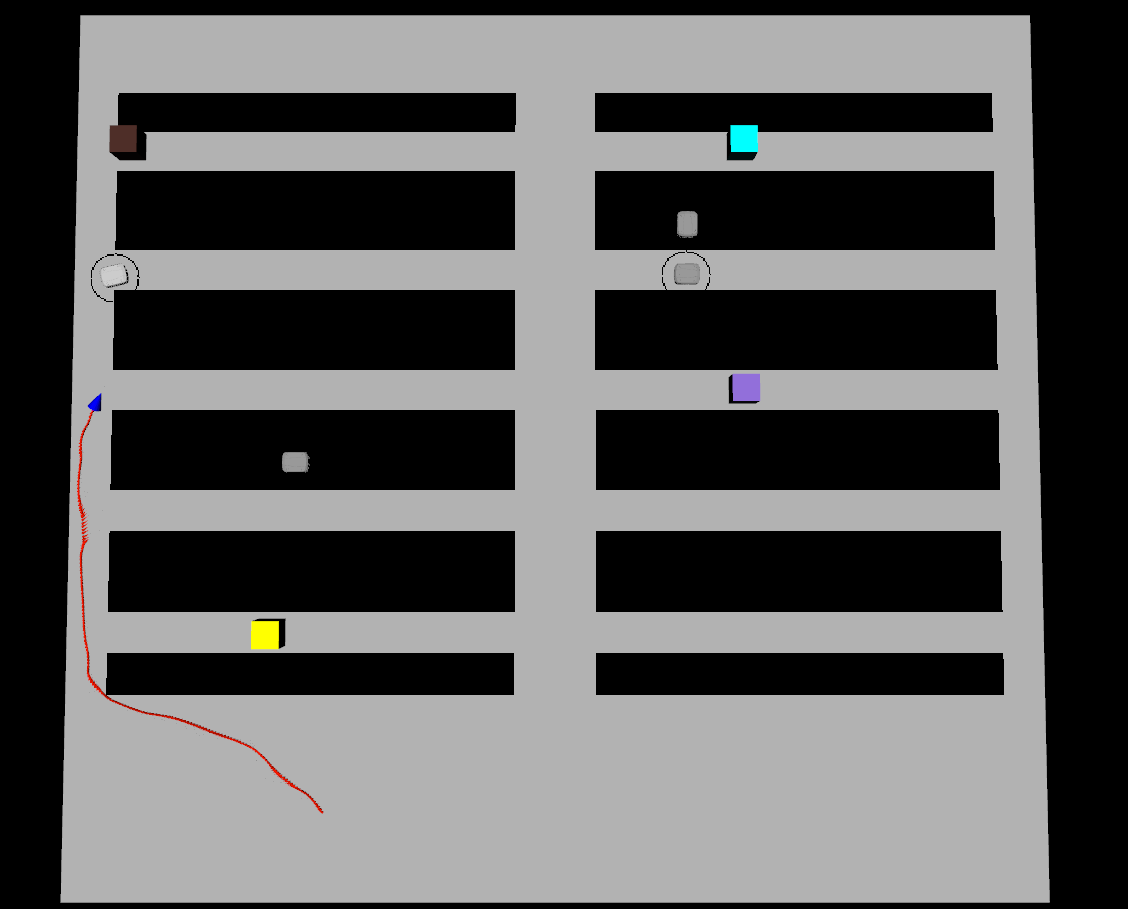}}\hspace*{.005\textwidth}
\subfloat[The worker has been moving towards the purple goal but has decided to turn left.]{\includegraphics[width= .33\textwidth]{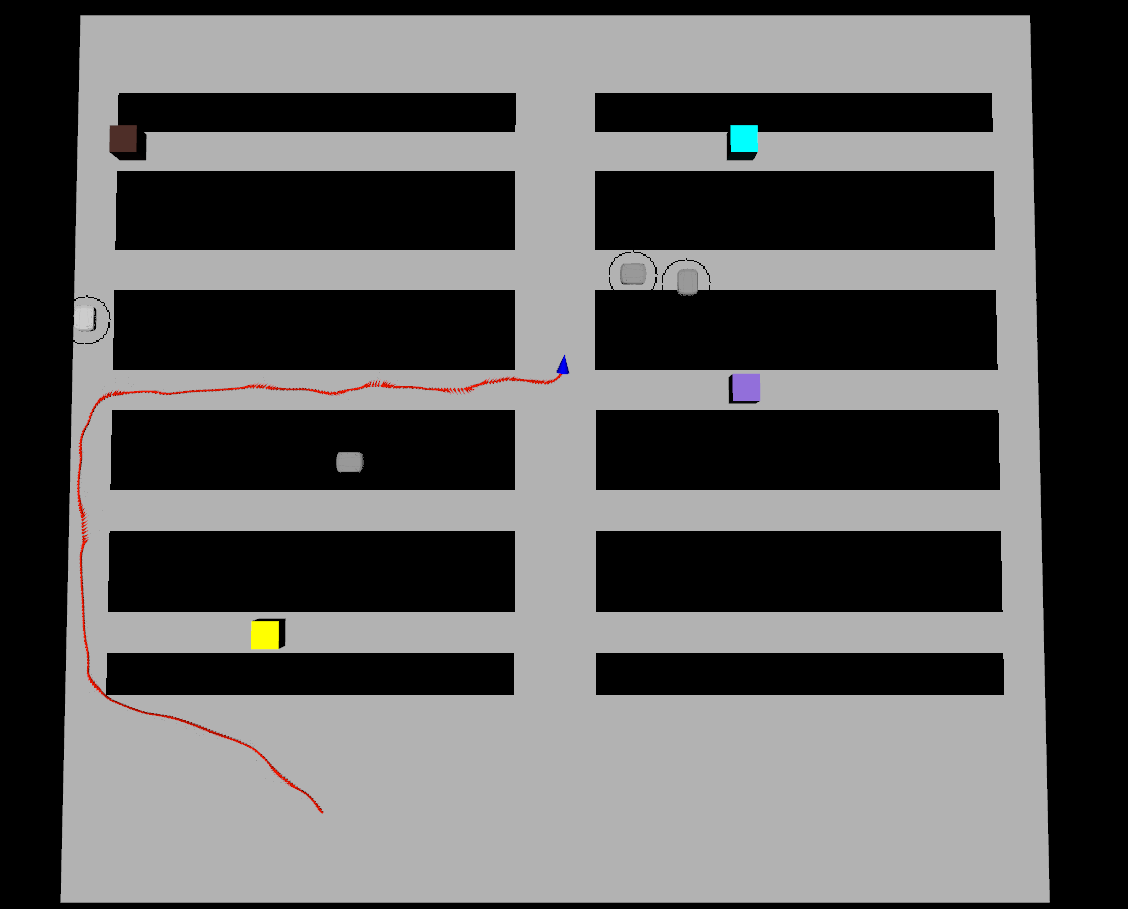}}
\vspace{-0.35cm}

\subfloat[The worker is on a crossroad. If it turns towards the brown goal it is obvious that it is the goal it wants to achieve.]{\includegraphics[width= .33\textwidth]{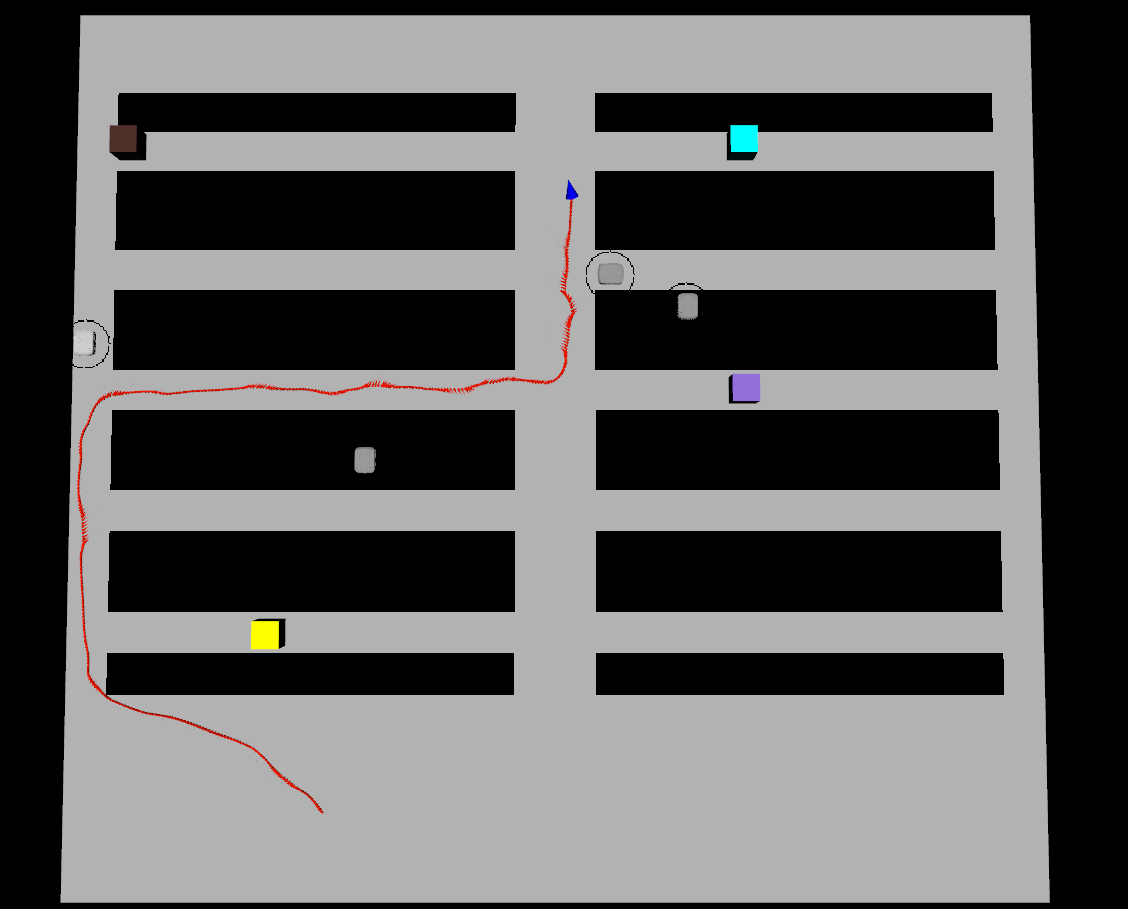}}\hspace*{.0015\textwidth}
\subfloat[The worker has turned towards the brown goal.]{\includegraphics[width= .33\textwidth]{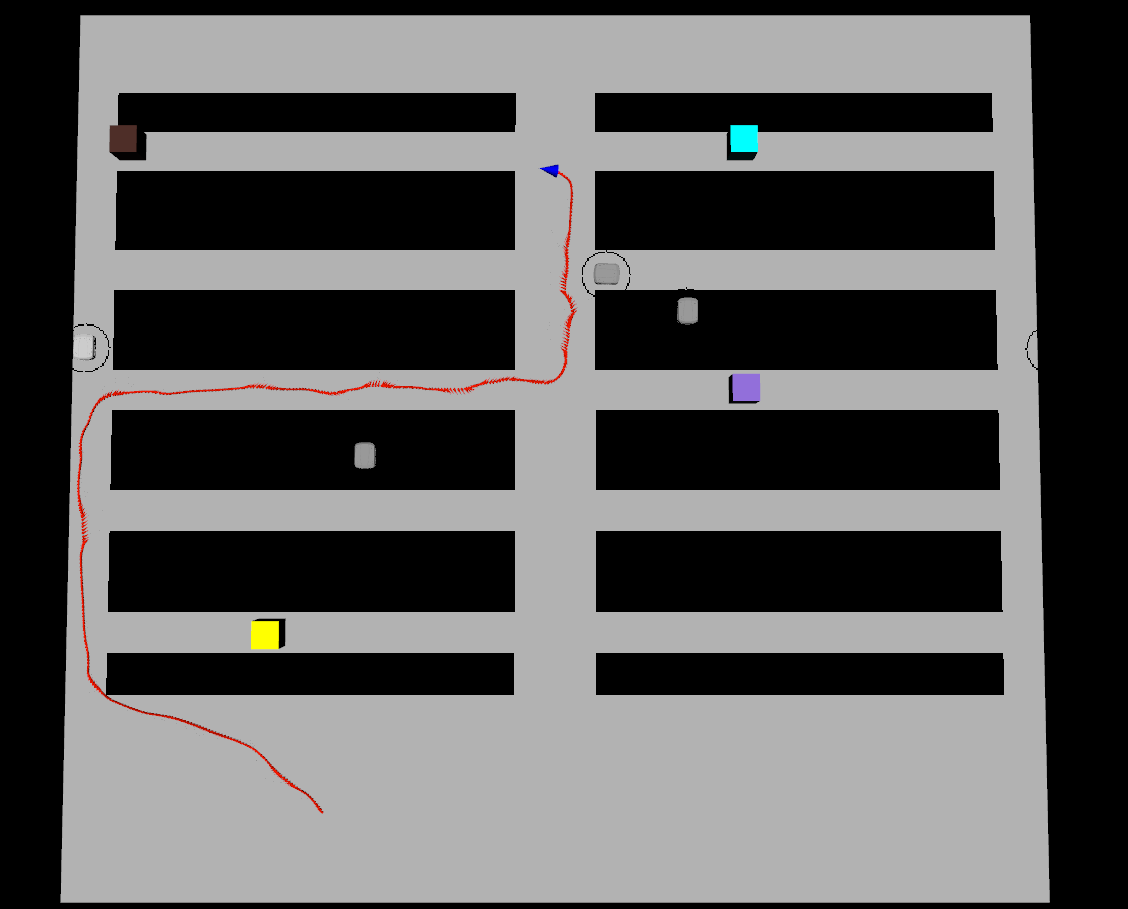}}\hspace*{.0015\textwidth}
\subfloat[End of the experiment.]{\includegraphics[width= .33\textwidth]{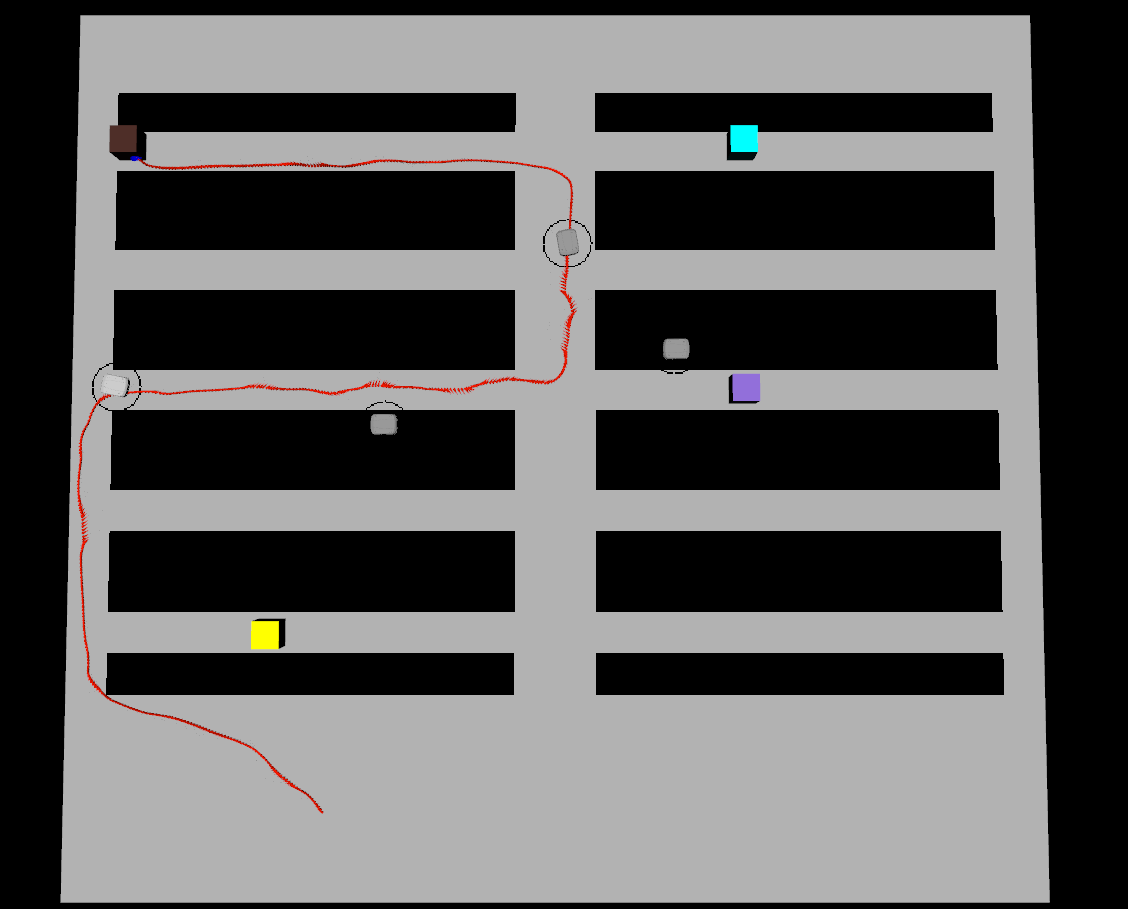}}
\vspace{-0.35cm}

\caption{Key moments of the virtual reality generated scenario using larger warehouse visualized in \emph{rviz}. Goal nodes are labeled with cubes and goal labeled with brown cube is being added during the experiment.}
\vspace{-0.35cm}

\label{fig:vr_results}
\end{figure*}

\begin{figure*}[!t]
\vspace{-0.8cm}
\centering
\subfloat[Algorithm's output with brown goal being known through the whole experiment.\label{VR_apriori}]{
  \includegraphics[width=.9\columnwidth]{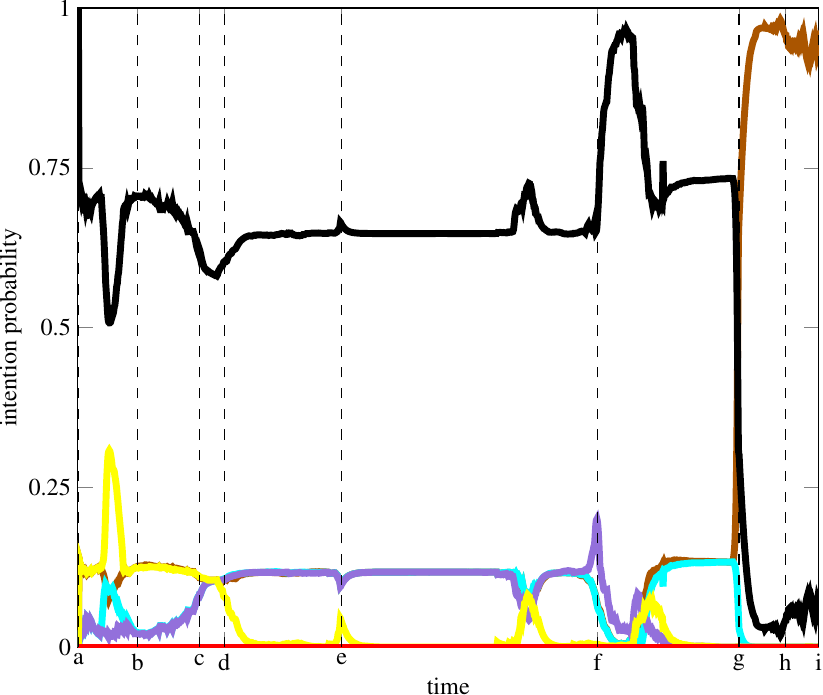}
  }
  \qquad
\subfloat[Algorithm's output with brown goal being added during the experiment.\label{VR_added}]{
  \includegraphics[width=.9\columnwidth]{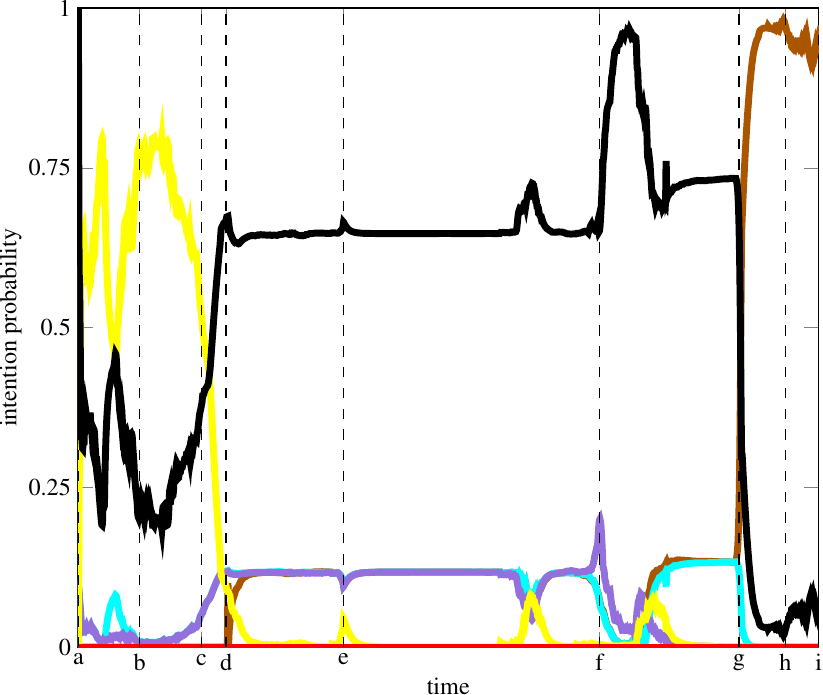}
  }
\caption{Comparison of the proposed algorithm's output in the larger VR warehouse between two different scenarios.
Intention estimations for three locations are labeled with respect to their color in Fig.~\ref{fig:vr_results} (brown, cyan and purple), the unknown goal state is labeled black and the irrational worker state is labeled red.
One can notice that the algorithm's outputs are different only on the segment where the brown goal has not been introduced in \ref{VR_added}.
}
\vspace{-0.35cm}

\label{probabilities_vr}
\end{figure*}

We then proceeded to conducting experiments in a larger VR warehouse (layout can be seen in Fig.~\ref{small_layout}) with four robots running on preprogrammed paths.
These paths were generated by the path planner and read from a text file containing JSON messages.
We considered two scenarios for this experiment: one with four initially known goals and one with three initially known goals with one goal being added during the experiment.
Visualization of the key moments can bee seen in Fig.~\ref{fig:vr_results} and the proposed algorithm output is shown in Fig~\ref{probabilities_vr}.
The worker initially starts moving towards yellow and brown goals.
If the brown goal is known, the model cannot decide which goal the worker desires more and stays in the unknown goal state (Fig.~\ref{VR_apriori}).
However, if the brown goal has not yet been added, worker moves only to the yellow goal and probability for that goal steeply rises (Fig.~\ref{VR_added}).
The worker then continues past the yellow goal which causes its intention estimation to fall.
Shortly after this event, the brown goal is added in second scenario.
Once the mobile robot blocks the shortest path towards the brown goal, the worker turns right.
Because it is now also moving towards cyan and purple goal, the model cannot decide which goal worker desires the most.
The worker continues moving towards the crossroad and hesitates with turning towards brown and cyan goals which manifests as a spike of estimation for purple goal.
However, since worker eventually turns towards brown and cyan goals, purple goal's estimation steeply falls.
The worker then continues moving towards the brown goal and the model recognizes its intention.

\begin{figure*}[htb]
\subfloat[Initial position.]{\includegraphics[width= .33\textwidth]{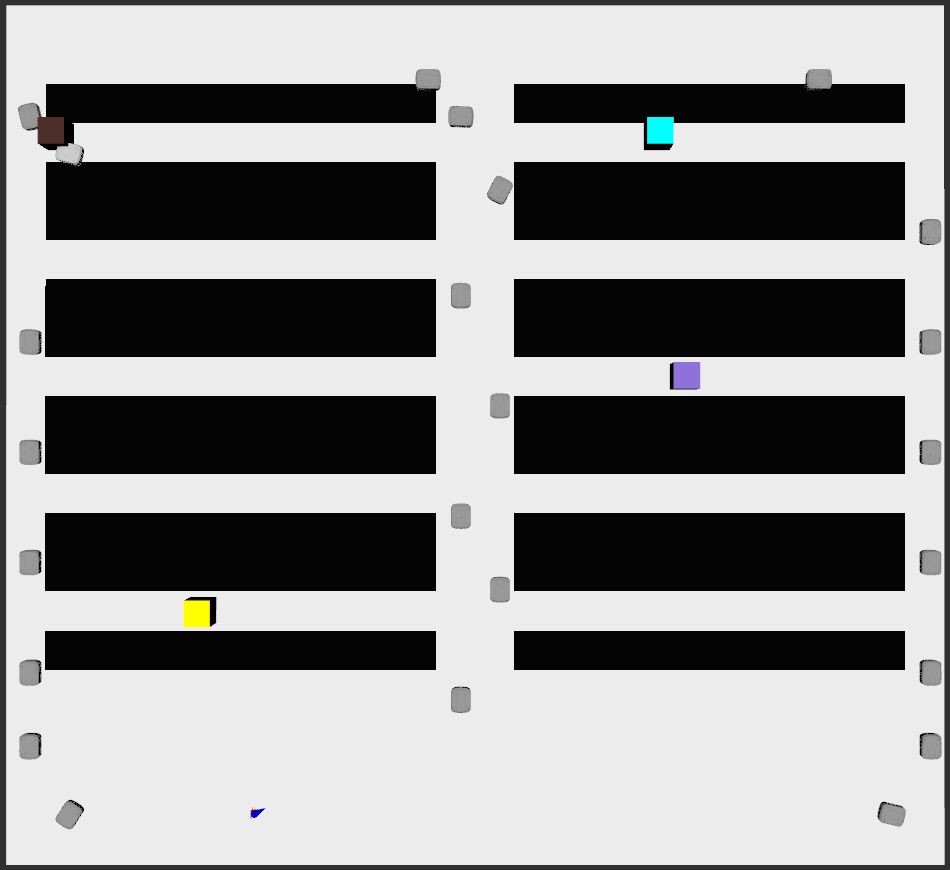}} \hspace*{.0015\textwidth}
\subfloat[The worker moves towards the yellow goal but because a mobile robot blocks the path towards it, the model declares the worker irrational.]{\includegraphics[width= .33\textwidth]{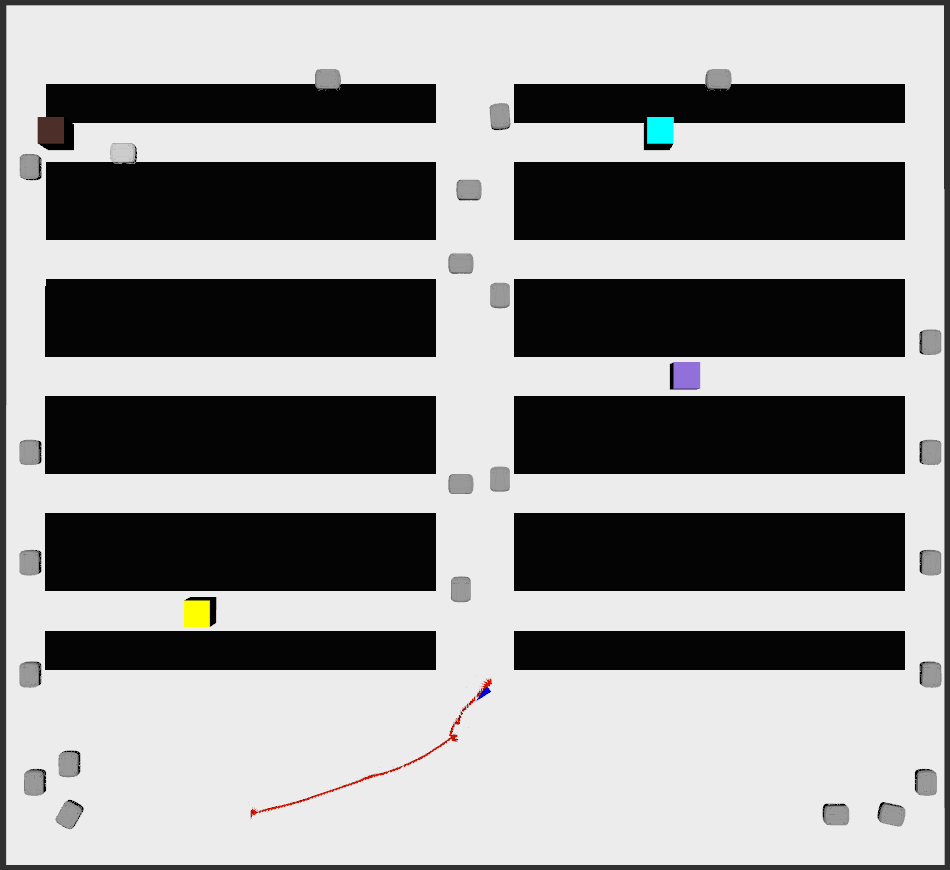}}\hspace*{.0015\textwidth}
\subfloat[Only the path towards the yellow goal is unobstucted but the worker does not go directly towards it.]{\includegraphics[width= .33\textwidth]{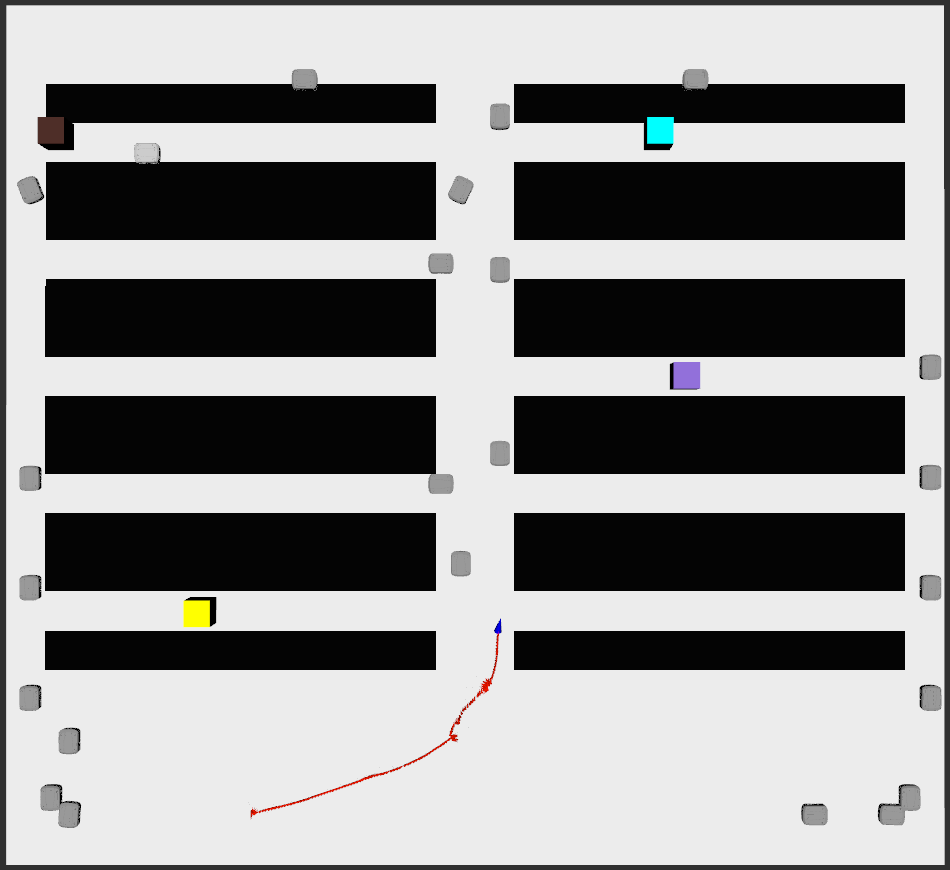}}
\vspace{-0.35cm}

\subfloat[The mobile robot blocks the worker's advancement. Because the worker turns and moves while waiting for the robot to move, the model switches estimation between the yellow goal and the unknown goal state.]{\includegraphics[width= .33\textwidth]{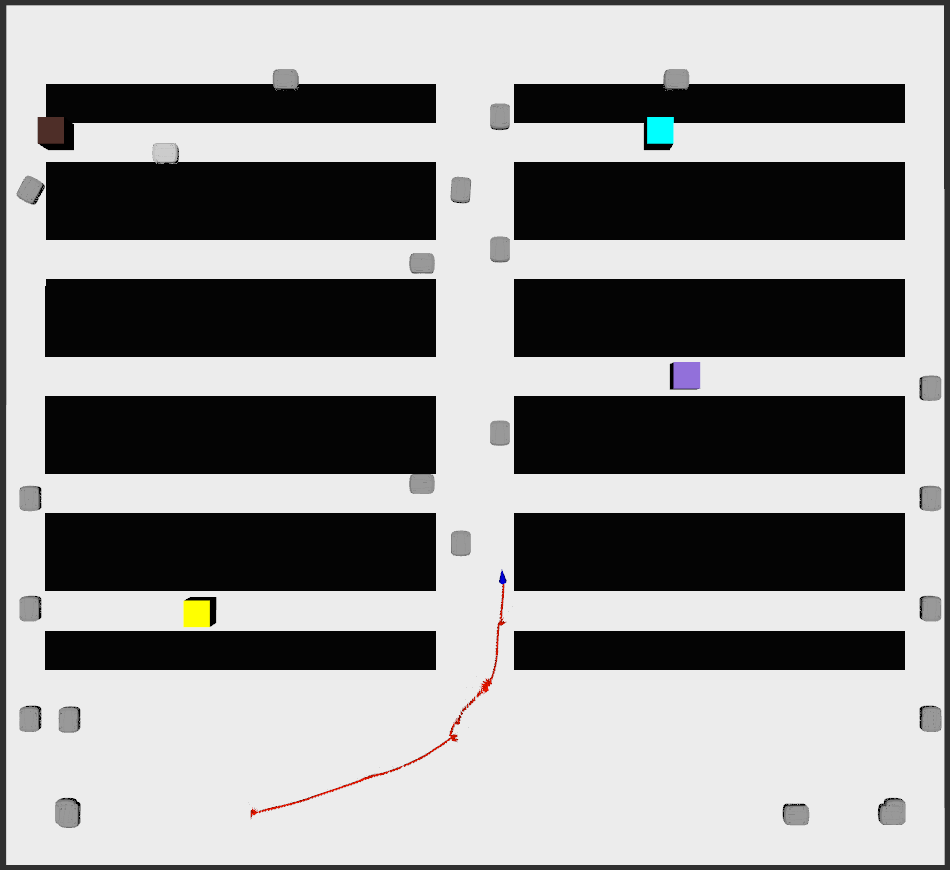}}\hspace*{.0015\textwidth}
\subfloat[Mobile robot moved and the worker is going to the purple goal.]{\includegraphics[width= .33\textwidth]{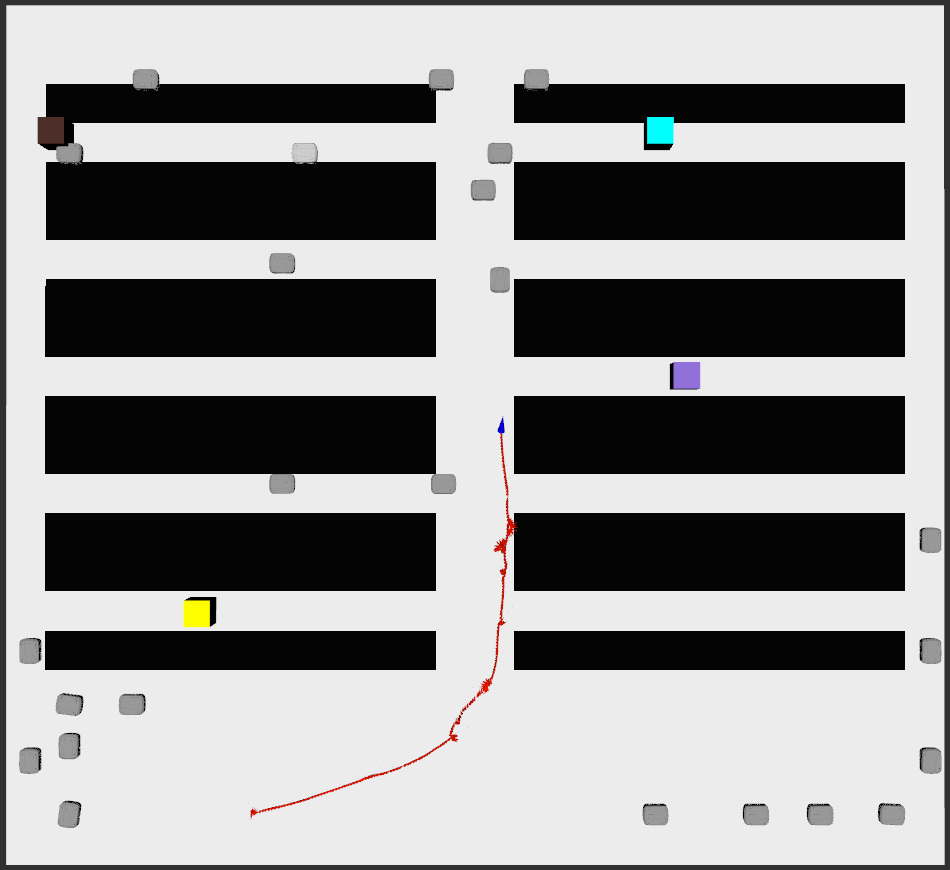}}\hspace*{.005\textwidth}
\subfloat[The worker is at the purple goal's location which ends the experiment.]{\includegraphics[width= .33\textwidth]{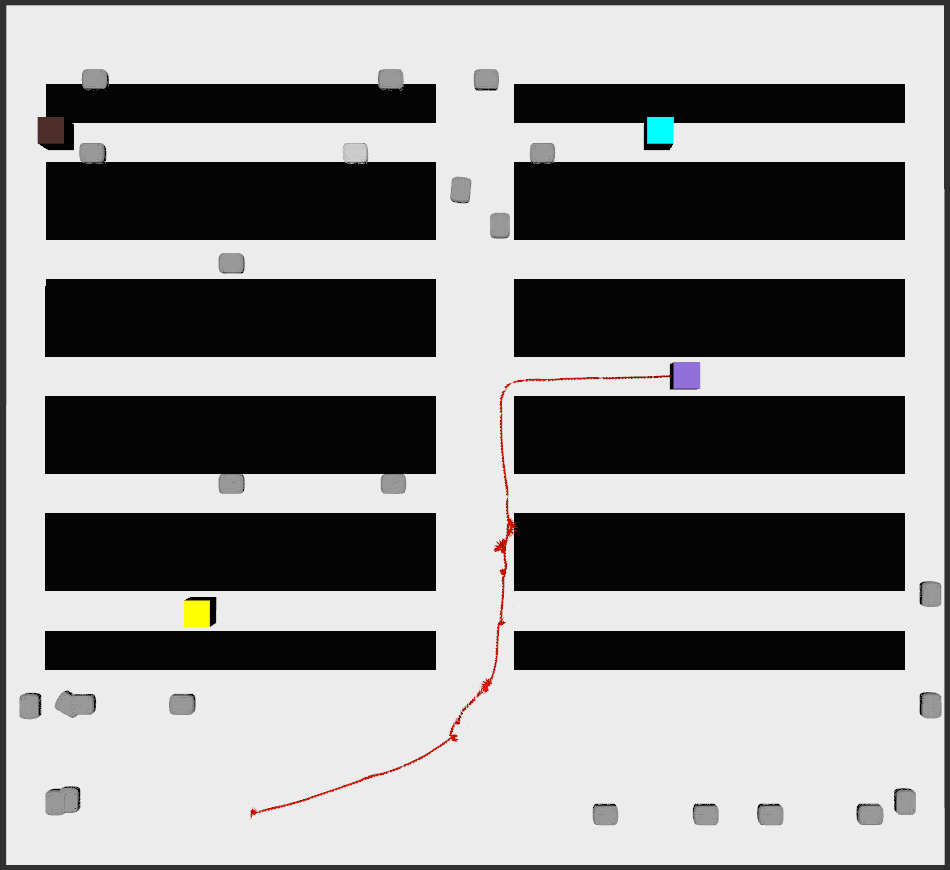}}
\caption{Key moments of the virtual reality generated scenario using a larger warehouse with 24 mobile robots visualized in \emph{rviz}. Goal nodes are labeled with colored cubes.}
\vspace{-0.35cm}

\label{fig:vr_24_results}
\end{figure*}

\begin{figure}[htb]
\centering
\includegraphics[width=.95\linewidth]{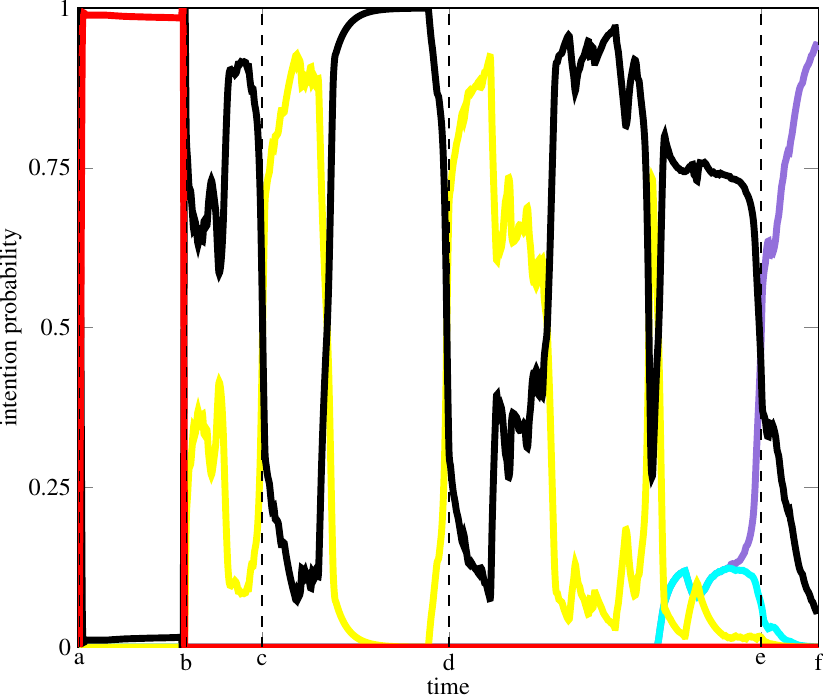}
\caption{Proposed algorithm's output in VR scenario with 24 mobile robots. Intention estimations for goal locations are labeled with respect to their color in Fig.~\ref{fig:vr_24_results}, the unknown goal state is labeled black and the irrational worker state is labeled red.}
\vspace{-0.35cm}

\label{probabilities_vr_24}
\end{figure}

In the end, we have conducted another experiment on the large-scale VR test warehouse with 24 mobile robots.
The robots were placed in groups of eight on the far left, middle and far right vertical corridors, equally spaced vertically.
Each robot then selected one of the reachable adjacent nodes at random and continued selecting nodes in such a manner, except that it is not allowed to return to the node visited in the previous step.
If another robot already selected the same node in the same time step, it stops and waits for it to pass.
This works well at emulating a fleet of robots moving around the warehouse without collisions.
The goal of this experiment was to show the proposed algorithm's scalability with the respect to the number of mobile robots.
We emphasize that mobile robot trajectories were not taking worker position into account.
Given that, in some scenes the worker is moving in the proximity of a mobile robot carrying a warehouse rack, which would not meet safety requirements in a realistic flexible robotized warehouses.
Nevertheless, we use this scenario solely for illustration purposes in order to demonstrate the proposed algorithm's scalability.
The key moments of the experiment are shown in Fig.~\ref{fig:vr_24_results} and the result of intention estimation is shown in Fig.~\ref{probabilities_vr_24}.


\section{Conclusion}
\label{sec:conclusion}
In this paper we have proposed a real-time human intention estimation algorithm capable of a dynamic environment.
Our goal was to estimate the intention of a human worker inside of a robotized warehouse whose layout can change due to robots blocking paths.
We assumed that the worker position and orientation are measured and that the warehouse layout and robots positions are readily available.
The worker has a set of potential goal locations that can be defined before or added during the experiment.
The task of the proposed algorithm is to precisely estimate worker's desires for each goal.
Given that, we evaluated worker actions with the respect to a modulated optimal path generated using the generalized Voronoi diagram and D$^*$ algorithm.
Then we used the resulting motion validation as observations of the hidden Markov model framework to estimate the final probabilities of worker intentions.
We have carried out multiple experiments both in a real-world industrial setup using augmented reality glasses and in virtual reality generated warehouses in order to demonstrate the scalability of the algorithm.
Results corroborate that the proposed framework estimates warehouse worker’s desires precisely and within reasonable expectations.

For future work, the worker intention estimation is planned to be used for robot human-aware trajectories replanning to ensure more efficient operation of flexible robotized warehouses.
Currently, the VR environment is used mostly for rapid prototyping and testing of AR interactions, while we are also looking into expanding it for training purposes and possibly system monitoring.
Furthermore, the AR interactions were developed for the \textit{Microsoft Hololens} and built upon the standard use of AR in logistics as part of \emph{pick-by-vision} systems and in the future it is envisaged to include navigational help, situational awareness and maintenance help for humans working in a mostly automated warehouse.

\section*{}
\begin{nomenclature}
\begin{deflist}[AAAA] 
\defitem{AR}\defterm{Augmented reality}
\defitem{BToM}\defterm{Bayesian Theory of Mind}
\defitem{FoV}\defterm{Field of view}
\defitem{GHMM}\defterm{Growing hidden Markov model}
\defitem{GVD}\defterm{Generalized Voronoi diagram}
\defitem{HMM}\defterm{Hidden Markov model}
\defitem{HRI}\defterm{Human robot interaction}
\defitem{HUD}\defterm{Heads-up display}
\defitem{MDP}\defterm{Markov decision process}
\defitem{MR}\defterm{Mixed reality}
\defitem{POMDP}\defterm{Partially observable Markov decision process}
\defitem{SLAM}\defterm{Simultaneous localization and mapping}
\defitem{VR}\defterm{Virtual reality}
\defitem{\textbf{F}}\defterm{Relative distances matrix}
\defitem{\textbf{I}}\defterm{Isolation matrix}
\defitem{\textbf{c}}\defterm{Association vector}
\defitem{\textbf{d}}\defterm{Approximate distance vector}
\defitem{\textbf{D}}\defterm{Approximate distance matrix}
\defitem{\textbf{v}}\defterm{Motion validation vector}
\defitem{$G_i$}\defterm{HMM hidden states}
\defitem{\textbf{T}}\defterm{HMM transition matrix}
\defitem{$\alpha, \beta, \gamma, \delta$}\defterm{Transition matrix parameters}
\defitem{\textbf{B}}\defterm{HMM emission matrix}
\defitem{$\Pi$}\defterm{HMM initial state}
\end{deflist}
\end{nomenclature}
\newpage
\section*{Acknowledgment}
This work has been supported from the European Union's Horizon 2020 research and innovation programme under grant
agreement No 688117 ``Safe human-robot interaction in logistic applications for highly flexible warehouses (SafeLog)'' and the DAAD/MZO grant ``Human localization and intention recognition based on wearable sensors''.
The authors would like to thank Julian Schanowski of the Karlsruhe Institute of Technology for his help with the virtual reality experiments.


\balance

\bibliographystyle{elsarticle-num}

\section*{References}
\bibliography{library}


%
%
%
\end{document}